\documentclass{article}

\usepackage[utf8]{inputenc} 
\usepackage[T1]{fontenc}    
\usepackage{hyperref}       
\usepackage{url}            
\usepackage{booktabs}       
\usepackage{amsfonts}       
\usepackage{nicefrac}       
\usepackage{microtype}      
\usepackage{xcolor}         
\usepackage{xspace}
\usepackage{caption} 
\usepackage{subcaption}
\usepackage{adjustbox}
\usepackage[round]{natbib}
\renewcommand{\bibname}{References}
\renewcommand{\bibsection}{\subsubsection*{\bibname}}

\usepackage{multirow}
\usepackage{multicol}

\usepackage{tikz}
\usepackage{pgfplots}
\usepgfplotslibrary{groupplots}
\pgfplotsset{compat=1.18}
\usetikzlibrary{external}

\usepackage{amsmath}
\usepackage{amssymb}
\usepackage{mathtools}
\usepackage{amsthm}
\usepackage{algorithm}
\usepackage{bm}

\usepackage{enumitem}

\usepackage[capitalize]{cleveref}

\theoremstyle{plain}
\newtheorem{theorem}{Theorem}[section]
\newtheorem{proposition}[theorem]{Proposition}
\newtheorem{lemma}[theorem]{Lemma}

\theoremstyle{definition}

\theoremstyle{remark}

\newcommand{\equal}{*}

\newcommand{\wrt}{w.r.t.\@\xspace}
\newcommand{\eg}{e.g.\@\xspace}

\newcommand{\iid}{i.i.d.\@\xspace}
\newcommand{\eqspace}{\qquad} 

\newcommand{\vek}[1]{\mathbf{#1}}

\renewcommand{\d}{\mathrm{d}}
\newcommand{\grad}[2][]{\nabla_{#1} #2 }
\newcommand{\expect}[2][]{\mathbb{E}_{#1} \left[ #2 \right]}

\newcommand{\argMin}[1]{\underset{#1}{\mathrm{argmin}}\, }
\newcommand{\klDiv}[2]{\mathrm{KL} \left[ #1 \| #2 \right] }
\newcommand{\eqDef}{\coloneqq}
\newcommand{\eqDefRight}{\eqqcolon}

\renewcommand{\neg}[1]{#1}
\newcommand{\unnorm}[1]{\tilde{#1}}
\newcommand{\param}{\theta}

\newcommand{\p}{p}
\newcommand{\pNoise}{q}
\newcommand{\pNoiseF}{\pNoise_{\varphi}}

\newcommand{\w}{\unnorm{w}_\theta}
\newcommand{\wabbrv}[1]{\w(\y_{#1})}
\newcommand{\wNormSpec}[1]{w} 
\newcommand{\wNorm}[1]{w_{#1}}
\newcommand{\wCond}[2]{\w(\y_{#1}\mid \y_{#2})}
\newcommand{\wNormCond}[2]{\wNorm{}{}_{#1\mid #2}}

\newcommand{\kernel}{K_{\param}}
\newcommand{\rankKernel}{\kernel^{\mathrm{R}}}
\newcommand{\condKernel}{\kernel^{\mathrm{C}}}

\newcommand{\pData}{\p_d}
\newcommand{\pModel}{\p_{\param}}
\newcommand{\pUnnorm}{\unnorm{\p}_{\param}}

\newcommand{\x}{\vek{x}}
\newcommand{\negY}{\neg{\vek{x}}}

\newcommand{\NumSamples}{N}
\newcommand{\numNeg}{j}
\newcommand{\NumNeg}{J}

\newcommand{\partFn}{Z_{\theta}}
\newcommand{\partFnHat}{\hat{Z}_{\theta}}
\newcommand{\partFnHatIs}{\partFnHat^{\text{IS}}}
\newcommand{\partFnHatCis}{\partFnHat^{\text{CIS}}}
\newcommand{\partFnHatCsmc}{\partFnHat^{\text{CSMC}}}

\newcommand{\Zz}{\partFn^z}

\newcommand{\gradP}[1]{\grad[\param]{#1}}
\newcommand{\gradF}{\grad[\varphi]{}}
\newcommand{\seq}[3]{ #1_{#2: #3} }
\newcommand{\seqNot}[2]{ #1_{- #2} }

\newcommand{\y}{\vek{x}}

\newcommand{\yNegSeq}[1]{ \seq{\negY}{#1}{\NumNeg}}
\newcommand{\ySeqNot}[1]{ \seqNot{\negY}{#1} }
\newcommand{\iter}[2]{ #1^{(#2)}}
\newcommand{\yIt}[1]{ \iter{\y}{#1}}

\newcommand{\bernoullivar}{z}
\newcommand{\selectInd}{z}
\newcommand{\totMcmcSteps}{k}
\newcommand{\mcmcStep}{\ell}
\newcommand{\epoch}{t}

\newcommand{\categorical}[1]{\text{Categorical} \left( #1 \right)}
\newcommand{\bernoulli}[1]{\text{Bernoulli}\left( #1 \right)}
\newcommand{\normal}[1]{ \mathcal{N} \left( {#1} \right)}

\newcommand{\lossFn}{\mathcal{L}}
\newcommand{\lossFnSub}[1]{\lossFn_{\text{#1}}}
\newcommand{\lossFnHat}{\hat{\mathcal{L}}}

\newcommand{\rankLossFn}{\lossFnSub{R}}
\newcommand{\condLossFn}{\lossFnSub{C}}

\newcommand{\twonorm}[1]{\|#1\|_2}
\newcommand{\identityMatrix}{\mathbb{I}}

\newcommand{\lr}{\kappa}
\newcommand{\batchSize}{B}
\newcommand{\xdim}{d}
\newcommand{\Xdim}{D}

\definecolor{color1}{RGB}{31,68,156}
\definecolor{color2}{RGB}{240,80,57}
\definecolor{color3}{RGB}{124,161,204}
\definecolor{color4}{RGB}{238,186,180}

\title{On the connection between Noise-Contrastive Estimation and Contrastive Divergence}

\newcommand{\authorliu}[2]{
{\normalsize{\bf #1}}\\
\normalsize{#2@liu.se} \\
\normalsize{Link\"oping University}
}

\newcommand{\authorchalmers}[2]{
{\normalsize{\bf #1}}\\
\normalsize{#2@chalmers.se} \\
\normalsize{Chalmers University of Technology}
}

\author{
\authorliu{Amanda Olmin$\footnote{Equal contribution}$}{amanda.olmin}
\and 
\authorchalmers{Jakob Lindqvist$^*$}{jakob.lindqvist}
\and
\authorchalmers{Lennart Svensson}{lennart.svensson}
\and
\authorliu{Fredrik Lindsten}{fredrik.lindsten}
}

\date{}

\begin{document}

\maketitle

\begin{abstract}
Noise-contrastive estimation (NCE) is a popular method for estimating unnormalised probabilistic models, such as energy-based models, which are effective for modelling complex data distributions. 
Unlike classical maximum likelihood (ML) estimation that relies on importance sampling (resulting in ML-IS) or MCMC (resulting in contrastive divergence, CD), NCE uses a proxy criterion to avoid the need for evaluating an often intractable normalisation constant.
Despite apparent conceptual differences, we show that two NCE criteria, ranking NCE (RNCE) and conditional NCE (CNCE), can be viewed as ML estimation methods. Specifically, RNCE is equivalent to ML estimation combined with conditional importance sampling, and both RNCE and CNCE are special cases of CD. These findings bridge the gap between the two method classes and allow us to apply techniques from the ML-IS and CD literature to NCE, offering several 
advantageous extensions. 
\end{abstract}
\section{Introduction}

Unnormalised probabilistic models, such as energy-based models \citep{ebm_og,how_to_train_ebm, flow_CE_of_EBM,improved_cd, implicit_cloning}, products of experts \citep{contr_div_hinton} and Markov random fields \citep{mrfs_score_matching},
can be used for modelling complex data distributions by trading exact normalisation for flexibility. 
Estimating unnormalised models is however not straightforward since maximum likelihood (ML) estimation involves the typically intractable normalisation constant.

One way to handle this challenge is to estimate the normalisation constant using importance sampling (IS), resulting in a learning algorithm denoted ML-IS.
In gradient-based learning, an alternative to ML-IS is contrastive divergence (CD) \citep{contr_div_hinton}, where Markov chain Monte Carlo (MCMC) sampling is used to 
approximate the gradient of the log-normalisation constant. 

A different solution to handling an intractable normalisation constant is to reformulate the model estimation as a binary classification problem, as done in noise-contrastive estimation (NCE) \citep{nce}.
In NCE, the model implicitly learns the data distribution by learning to distinguish between true samples and samples from a \textit{noise} distribution.

Several extensions of 
NCE have been proposed: mainly ranking NCE (RNCE), which is a multi-class version of its predecessor \citep{nce_ranking_obj}, and conditional NCE (CNCE), where the noise distribution is conditioned on the data \citep{cond_nce}. RNCE in turn, has been extended into new estimation methods \citep{how_to_train_ebm, nce_neg_sampl_cond_models}.
In \citet{flow_CE_of_EBM} a version of NCE is proposed, where the data and noise distributions are jointly learned.

Evidently, there is a plethora of methods for estimating unnormalised models, some of which seem conceptually different.
We hence argue for a need to create a more coherent framework. To contribute to this objective, we 
provide a direct relationship between NCE and ML-IS as well as CD.
We believe that this link makes it easier to understand and analyse the methods, and brings additional theoretical insights apart from what has previously been established \citep{nce_neg_sampl_cond_models, cond_nce}.
Specifically, we strengthen the connection between NCE, ML-IS and CD by:
%
\begin{itemize}[leftmargin=10pt,topsep=0em,itemsep=0.01em]
    \item clarifying the connection between RNCE and standard IS by showing that RNCE can be derived through an extension of IS, referred to as conditional IS (CIS) \citep{pmcmc,elements_of_smc};
    \item  showing that both CNCE and RNCE are special cases of CD, with kernels based on CIS and a variant of the Metropolis--Hastings algorithm, respectively.
\end{itemize}

Previous work has connected the original NCE criterion to general (statistical) frameworks based on Bregman divergences \citep{bregman_div_est_unnorm, unified_est_framework_unnorm_models}, but to the best of our knowledge, RNCE and CNCE have not been connected to such frameworks. RNCE has previously been linked to IS \citep{nce_ranking_obj},
but in an informal way and without making the connection to CIS, which is necessary for the equivalence to hold.
Moreover, another common method for training unnormalised models, namely, score matching, is a limiting case of CNCE \citep{cond_nce}. This is especially interesting as also CD has links to score matching \citep{connection_cd_sm}. 

Closest to our contribution is the work by \citet{cd_time_rev}. They focus on CNCE and show that CD is recovered by selecting the noise distribution in CNCE as an MCMC kernel.
We go in the other direction and show that CNCE is in fact a special case of CD \emph{for any noise distribution}. Importantly, this interpretation holds in the general case, without the need of re-weighting the CD gradient, as done in \citep{cd_time_rev}.
Furthermore, for multi-step sampling we can rely on standard MCMC theory and do not need to introduce the "time-reversal adversarial game" as proposed by \citet{cd_time_rev}. Finally, we consider also RNCE, which they do not.

Based on the established connections, we use techniques from existing literature on ML-IS and CD to 
(i) theoretically justify why RNCE empirically performs better than ML-IS \citep{how_to_train_ebm}, (ii) motivate why, for optimal learning, the noise distribution should resemble the model distribution, and not the data distribution as proposed previously \citep{nce} and (iii) identify several extensions to RNCE and CNCE and 
show empirically that these improve performance, with little or no increase in computational cost.  
We hope that these connections can lead to more valuable insights, apart from those identified in this paper.
\section{Background}
Given \iid training data $\{\x_0^i\}_{i=1}^N$ from some unknown data distribution $\pData(\cdot)$, we seek to approximate $\pData(\cdot)$  with a parametric model
\begin{align}
    \label{eq:unnorm_model}
    \textstyle
    \pModel(\y) =  \pUnnorm(\y) / \partFn,
    \quad \partFn = \int \pUnnorm(\y') \d \y',
\end{align}
where $\pUnnorm$ is the unnormalised model and $\partFn$ is the normalisation constant.
We assume that $\pModel \gg \pData$, meaning that $\pModel$ covers the support of $\pData$, such that $\pModel(\x) > 0$ whenever $\pData(\x) > 0$.
The model is estimated by minimising some criterion $\lossFn(\param)$ with respect to (\wrt) the parameter vector $\param$.
For the ML estimator, the criterion is the negative log-likelihood (NLL) of the model
\begin{align}
    \label{eq:unnorm_model:ll}
    \lossFn(\param; \y_0) = - \log \pModel(\x_0)
    &= - \log \pUnnorm(\y_0) + \log \partFn.
\end{align}
We use $\y_0$ to denote a sample from $\pData$.
In practice, $\lossFn(\param; \y_0)$ is computed as an average over $\NumSamples$ independent samples $\x_0^i \sim \pData(\cdot)$ but for brevity we perform all derivations for a single data point $\y_0$.

The first term in \cref{eq:unnorm_model:ll} is normally easy to evaluate but the second term involves the typically intractable integral in \cref{eq:unnorm_model}.
Below, we introduce common methods for handling this.
Derivations and proofs that are omitted from the main article can be found in the supplementary material.

\subsection{Importance sampling}
Using an importance sampling (IS) estimate of  $\partFn$ in \cref{eq:unnorm_model:ll} results in an approximate ML criterion abbreviated ML-IS.
Assume that we have defined a proposal distribution $\pNoise$, such that $\pNoise \gg \pModel$.
In IS, we draw $\NumNeg$ i.i.d. samples 
$\yNegSeq{1} = [\y_1, \dots, \y_\NumNeg]$ from $\pNoise$; we use $\pNoise(\yNegSeq{1}) \eqDef \prod_{j=1}^\NumNeg \pNoise(\y_j)$ to denote their joint distribution.
Given $\yNegSeq{1}$, we approximate the normalisation as
\begin{align}
    \label{eq:is:z_hat}
    \partFn &\approx \partFnHatIs \eqDef \frac{1}{\NumNeg} \sum_{\numNeg=1}^{\NumNeg} \wabbrv{\numNeg},
    \, \text{with}\,
    \wabbrv{\numNeg} \eqDef \frac{\pUnnorm(\y_{\numNeg})}{\pNoise(\y_{\numNeg})}.
\end{align}
The estimate $\partFnHatIs$ is unbiased, see e.g. \citep{elements_of_smc}, but the gradient $\gradP \log \partFnHatIs$ is not \citep{mc_stat_met}. Meanwhile, having an unbiased gradient estimate is undoubtedly advantageous. It is, for example, a standard condition for proving general convergence of stochastic gradient descent (SGD) \citep{optim_large_scale}.

\subsection{Contrastive divergence}
Instead of approximating the NLL in \cref{eq:unnorm_model:ll}, we can estimate its gradients using the identity \citep{ebm_og}%
\begin{align}
    \label{eq:unnorm_model:grad}
    &-\gradP{} \log \pModel(\y_0) = 
   -\gradP{\log \pUnnorm(\y_0)}  + \expect[\pModel(\y')]{ \gradP{\log \pUnnorm(\y')} }.
\end{align}
Again, this gradient is intractable since it involves an expected value \wrt $\pModel$,
but it can be approximated with MCMC methods.
Contrastive divergence (CD) \citep{contr_div_hinton}, uses the approximation
\begin{align}
    \label{eq:cd:grad}
    &-\gradP{ \log \pModel(\y_0) } \approx &- \gradP{\log \pUnnorm(\y_0) + \expect[\kernel(\y' \mid \y_0) ]{ \gradP{\log \pUnnorm(\y')} } },
\end{align}
where $\kernel(\y' \mid \y_0)$ is a $\pModel$-invariant MCMC-kernel initialised at a data sample $\y_0$.

For efficiency, CD runs the MCMC chain for only a few steps or even a single step (CD-1). This is typically not enough to obtain a sample from the target distribution $\pModel$. However, initialising the chain with a sample from the data distribution is a means of initialising it close to the modes of $\pModel$, especially as $\pModel$ converges to $\pData$.
Empirical evidence suggests that the bias of the CD parameter estimate, in relation to the ML estimate, is small \citep{on_cd_learning}. 

CD was originally based on a different criterion than the log-likelihood \citep{contr_div_hinton}.
However, it is common to derive CD as done here, see, e.g., \citep{cd_wormholes,mcmc_mle}, and the resulting gradient expression is the same.

\subsection{Noise-contrastive estimation}
Noise-contrastive estimation (NCE) avoids computing the NLL in \cref{eq:unnorm_model:ll} altogether, 
by transforming the model estimation into a task of discriminating between true samples from $\pData$ and samples from a noise distribution (Gutmann and Hyvärinen, 2012; Pihlaja et al., 2010).
For later comparison, we interpret the noise distribution as a proposal and denote it with $\pNoise$.
We focus on two extensions of the original NCE method:
RNCE \citep{nce_ranking_obj} and CNCE \citep{cond_nce}. The first is consistent under weaker assumptions than standard NCE \citep{nce_neg_sampl_cond_models}, while the other improves upon the original formulation by conditioning the noise distribution on the data. Furthermore, both eliminate the need to include the normalisation constant as an extra model parameter.

Ranking NCE (RNCE), extends the binary classification problem of original NCE to a multi-class classification problem \citep{nce_ranking_obj}.
Consider a scenario with one data sample $\y_0 \sim \pData(\cdot)$ and $\NumNeg$ i.i.d. noisy samples $\yNegSeq{1} \sim \pNoise(\cdot)$. 
As in ordinary NCE, we train the model to classify $\y_0$ as the sample coming from $\pData$.
Specifically, RNCE maximises the posterior distribution
\(
    p(\selectInd=0 \mid \y_{0:\NumNeg}),
\)
where $\selectInd \in \{0,1,\dots,\NumNeg\}$ is a latent categorical variable corresponding to the index, or class, of the data point drawn from $\pData$.
When calculating the posterior, the unknown data distribution $\pData$ is replaced with $\pModel$, forcing $\pModel$ to approach $\pData$.
With a uniform prior on $\selectInd$, the  empirical loss for one data point $\y_0\sim \pData(\cdot)$ becomes
\begin{align}
    \textstyle
    \label{eq:rank_nce:crit}
    \rankLossFn(\param; \yNegSeq{0})
    &= - \log \wabbrv{0} + \log \bigg( \sum_{\numNeg=0}^{\NumNeg} \wabbrv{\numNeg} \bigg),
\end{align}
where we re-write the criterion with the weights from \cref{eq:is:z_hat}; 
see \cref{app:rnce-criterion} for the derivation.

Another extension is Conditional NCE (CNCE) \citep{cond_nce} which allows the noise distribution to depend on the data sample, resulting in more difficult discrimination and thereby better model estimation.
CNCE, like standard NCE, considers a binary classification problem, where a data point $\y_0 \sim \pData(\cdot)$ is discriminated from a sample $\y_1 \sim \pNoise(\cdot \mid \y_0)$.

Following \cite{cond_nce}, we average the posterior of the latent (Bernoulli) class variable $\selectInd \in \{0, 1\}$ over $\NumNeg$ i.i.d. samples $\x_{\numNeg} \sim \pNoise(\cdot \mid \y_0)$, to reduce the variance of the estimate, and minimise
\begin{align}
    \label{eq:cond_nce:crit}
  \condLossFn(\param; \yNegSeq{0})
   & = \frac{1}{\NumNeg} \sum_{\numNeg=1}^{\NumNeg} \log \left( 1 + \frac{\wCond{\numNeg}{0}}{\wCond{0}{\numNeg}} \right). 
\end{align}
See \cref{app:cnce-criterion} for the derivation. The weights from \cref{eq:is:z_hat} are modified as
\begin{align}
    \label{eq:cnce:weight}
    &\wCond{\ell}{k} = \frac{\pUnnorm(\y_{\ell}) }{ \pNoise(\y_{\ell} \mid \y_{k})},
    \quad \ell, k \in \{0, \ldots, \NumNeg\}.
\end{align}
For brevity, we omit the dependency on $\y_0, \yNegSeq{1}$ in the criteria from here on.

\section{Importance sampling and RNCE}
In this section, we explore the connection between RNCE and ML-IS. 
Conceptually, ML-IS and RNCE are two distinct methods for estimating unnormalised models, but their gradient updates are very similar.
Indeed, \citet{nce_ranking_obj} claim that with the RNCE criterion they
``derive a similar surrogate classification task akin to NCE which arrives at IS''.
We show here that this statement is not entirely accurate as there is a slight, but important, difference.

The gradient of the RNCE criterion (\cref{eq:rank_nce:crit}) is 
\begin{align}
  \label{eq:rank_nce:grad}
  \gradP{} \rankLossFn(\param) &= - \gradP{} \log \pUnnorm(\y_0) + \sum_{\numNeg=0}^{\NumNeg} \wNorm{\numNeg} \gradP{} \log \pUnnorm(\y_\numNeg), \nonumber \\ 
  \wNorm{\numNeg} &= \frac{ \wabbrv{\numNeg} }{ \sum_{\ell=0}^{\NumNeg} \wabbrv{\ell} },
\end{align}
where $\wNorm{\numNeg}$ is the weight of the $j$th sample, normalised over all samples $\yNegSeq{0}$.
There is a subtle difference between this gradient and the ML-IS gradient.
For RNCE, the data sample $\y_0 \sim \pData(\cdot)$ is included in the sum in the second term.
However, for ML-IS, the second term, 
corresponding to the estimate of $\gradP{} \log \partFn$, only includes samples from the proposal distribution $\pNoise$.

Instead, we show that RNCE corresponds to an ML criterion based on conditional importance sampling (CIS).
CIS is a special case of Particle MCMC \citep{pmcmc,elements_of_smc}.
It is almost identical to standard IS, except that we condition on a sample $\y_0$.
For our task, the CIS estimator is defined as
\begin{equation}
    \label{eq:cis:z_hat}
    \partFn \approx \partFnHatCis \eqDef \frac{1}{\NumNeg+1} \sum_{\numNeg=0}^{\NumNeg} \wabbrv{\numNeg}.
\end{equation}
where $\yNegSeq{1} \sim \pNoise(\cdot)$ and $\y_0$ is the conditioned sample.

Assuming that $\y_0 \sim \pData(\cdot)$, we can derive the RNCE criterion directly from CIS.
\begin{proposition}[RNCE is ML-CIS]
\label{thm:rank_nce:cis} 
RNCE is equivalent to ML estimation using CIS, conditioning on $\y_0 \sim \pData(\cdot)$, for estimating the normalisation constant in \cref{eq:unnorm_model:ll}.
\end{proposition}
See \cref{app:proof-rnce-cis} for the proof.

This link between RNCE and ML-CIS unifies these seemingly different methods of estimation, offering us a deeper understanding of RNCE and allowing us to reason about the (relative) empirical performance of this method. Interestingly, in the special case where $\pModel = \pData$, such that $\y_0$ is a sample from the model distribution, the following holds: 
\begin{proposition}[Unbiased CIS estimate of $\gradP{} \log \partFn$]
    \label{thm:unbiased_cis:log_part_fn}
    Assume $\y_0 \sim \pModel(\cdot)$, $\yNegSeq{1} \sim \pNoise(\cdot)$ and that $\pNoise$ is independent of $\param$.
    Then, the CIS estimator gives an unbiased estimate of the gradient of the log-normalisation constant
    \begin{align}
        \expect[\pModel(\y_0), \pNoise(\yNegSeq{1})]{\gradP{} \log \partFnHatCis} = \gradP{} \log \partFn.
    \end{align}
\end{proposition}
See \cref{app:proof:prop-unbiased-cis} for the proof.

Note that, in practice, the conditions of \cref{thm:unbiased_cis:log_part_fn} are not strictly fulfilled for RNCE, since it uses the ``approximation'' $\y_0 \sim \pData(\cdot)$.
However, as $\pModel$ aims to approximate $\pData$, we hope that $\pModel \approx \pData$, at least during the later stages of training.
If the data distribution is a good substitute for the model distribution, we then obtain ``approximately unbiased'' estimates of $\gradP{} \log \partFn$.
By conditioning on the extra sample $\y_0$ we therefore get an advantage over IS, which does not give unbiased gradient estimates \citep{mc_stat_met}.
We hypothesise that this property of CIS can help explain why RNCE appears to give better gradient estimates, compared to ML-IS.
This is further motivated by the fact that the bias of RNCE will decrease as $\pModel$ converges to $\pData$ while the bias of IS will not.
\section{Connecting NCE with CD}
\label{sec:nce_cd_cis}
In this section, we connect RNCE and CNCE to the family of CD methods, and specifically CD-1, i.e.,
CD where the expectation in \cref{eq:cd:grad} is approximated with a single MCMC step.
It has been shown that CD-1 is a special case of CNCE if $\pNoise$ is an MCMC kernel fulfilling detailed balance \citep{cd_time_rev}.
Here, we show the reverse: that not only CNCE but also RNCE are special cases of CD-1.
The idea is to construct $\pModel$-invariant kernels, such that the gradient estimates of the resulting CD-1 variants are equivalent to those of RNCE and CNCE respectively.
We note that the assumptions that we make are standard in the NCE literature, and therefore that the equivalences hold whenever the NCE methods are applicable.

\subsection{RNCE criterion}
To show that RNCE is a CD-1 method, we introduce an MCMC kernel 
$\rankKernel(\y' \vert \y_0)$
based on CIS. \Cref{alg:cd_rnce:rnce_kernel} shows how to generate a sample from this kernel conditioned on the state $\x_0$.
We note that CIS was initially introduced as an MCMC procedure and that the kernel $\rankKernel(\y' \vert \y_0)$ is known to be $\pModel$-invariant, see e.g. \citep{pmcmc,elements_of_smc}. It thus fits into the CD framework.

\begin{figure*}[!t]
\begin{minipage}{0.45\textwidth}
\begin{algorithm}[H]
    \caption{CIS kernel}
    \textbf{Input:} $\x_0$
    \begin{enumerate}[topsep=0.2em, itemsep=0.05em]
        \item Sample $\y_{1:\NumNeg} \sim \pNoise(\cdot)$
        \item Calculate weights $\wNorm{\numNeg}$, $\numNeg = 0, \ldots, \NumNeg$, using \cref{eq:is:z_hat,eq:rank_nce:grad}
        \item Sample $\selectInd \sim \categorical{[\wNorm{0}, \ldots, \wNorm{\NumNeg}]}$
        \item Return $\y' = \y_{\selectInd}$
    \end{enumerate}
    \label{alg:cd_rnce:rnce_kernel}
\end{algorithm}
\end{minipage}\hspace*{0.7cm}
\begin{minipage}{0.45\textwidth}
\begin{algorithm}[H]
    \caption{CNCE kernel}
    \textbf{Input:} $\y_0$
    \begin{enumerate}[topsep=0.2em, itemsep=0.05em]
        \item Sample $\y_1 \sim \pNoise(\cdot \vert \y_0)$
            \item Calculate the weight $\wNormCond{1}{0}$, using \cref{eq:cnce:weight,eq:cond_nce:cnce_w_norm}
        \item Sample $\bernoullivar \sim  \bernoulli{\wNormCond{1}{0}}$
        \item Return $\y' = \y_{\bernoullivar}$
    \end{enumerate}
    \label{alg:cd_cnce:cnce_kernel}
\end{algorithm}
\end{minipage}
\end{figure*}

Using $\rankKernel(\y' \vert \y_0)$ in CD-1 to estimate the expectation in \cref{eq:cd:grad},
we can exactly recover the gradient of the RNCE criterion in \cref{eq:rank_nce:grad}.
This connection is formalised in \cref{prop:cd_rnce:cd_connection}.

\begin{proposition}[RNCE = CD-1]
    Model estimation with the RNCE criterion (\cref{eq:rank_nce:crit}) is a special case of CD-1,
    using the MCMC kernel $\rankKernel(\y' \vert \y_0)$ defined in \cref{alg:cd_rnce:rnce_kernel} if, when evaluating the expected value in \cref{eq:cd:grad}, the variable $\selectInd$ 
    used in \cref{alg:cd_rnce:rnce_kernel}
    is marginalised out.\label{prop:cd_rnce:cd_connection}
\end{proposition}
See \cref{app:proof-cd-rnce} for the proof.

\subsection{CNCE criterion}
Next, we establish a connection between CNCE and CD.
First, the gradient of the CNCE criterion in \cref{eq:cond_nce:crit} can be written as
(see \cref{app:cnce-gradient})
\begin{align}
    \label{eq:cond_nce:grad}
     \gradP{} \condLossFn(\param) 
    &=   -\gradP{}\log \pUnnorm(\y_0) 
     + \frac{1}{\NumNeg} \sum_{\numNeg=1}^{\NumNeg} \Big( 
     (1-\wNorm{}{}_{\numNeg\mid 0}) \nonumber \\ \eqspace & \cdot \gradP{} \log \pUnnorm(\y_0)
     + \wNorm{}{}_{\numNeg \mid 0} \gradP{} \log \pUnnorm(\y_{\numNeg}) \Big), 
    \\   \wNormCond{j}{0} &= \frac{\w(\y_{\numNeg} \mid \y_{0})}{\w(\y_{\numNeg} \mid \y_{0}) + \w(\y_{0} \mid \y_{\numNeg})}.
    \label{eq:cond_nce:cnce_w_norm}
\end{align}

With the CD framework, we can derive the CNCE criterion by formulating a kernel $\condKernel(\x'\mid \x_0)$, conditioned on $\y_0$, according to \cref{alg:cd_cnce:cnce_kernel}. The kernel is similar to the Metropolis--Hastings algorithm \citep{mh_metropolis, mh_hastings}, but uses another acceptance probability, which was also considered by \citet{mh_hastings}.
For a symmetric proposal distribution, i.e. $\pNoise(\y_1 \vert \y_0) = \pNoise(\y_0 \vert \y_1)$, it reduces  to Barker's method \citep{barkers_method}.
The kernel is $\pModel$-invariant as it fulfils the detailed balance condition, see \citep{mh_hastings}.

The main result concerning CNCE and its connection to CD, is given by \cref{prop:cd_cnce:cd_connection}.
\begin{proposition}[CNCE = CD-1]
   Model estimation with the CNCE criterion (\cref{eq:rank_nce:crit}) is a special case of CD-1, using the MCMC kernel defined in \cref{alg:cd_cnce:cnce_kernel} if, when estimating the expected value in \cref{eq:cd:grad}:
       \textit{(i)} an average is taken over $\NumNeg$ independent samples $\y_{\numNeg} \sim \condKernel(\cdot \vert \y_0)$, and \textit{(ii)} the variable $\bernoullivar$ used in \cref{alg:cd_cnce:cnce_kernel}
       is marginalised out for each sample.
    \label{prop:cd_cnce:cd_connection}
\end{proposition}
See \cref{app:proof-cd-cnce} for the proof.

In contrast to \cite{cd_time_rev}, where CD-1 is derived from CNCE using the specific choice of $\pNoise$ as the CD kernel, we derive CNCE from CD-1, for any choice of $\pNoise$.
While \cite{cd_time_rev} do consider general $\pNoise$, it is viewed as an extension of CD-1, where the gradient is by design re-weighted to match that of the CNCE criterion.
In our derivation, this re-weighting instead follows naturally from a Rao-Blackwellisation of the MCMC kernel, i.e. a marginalisation of the latent variable $\bernoullivar$, and no additional weighting is required to recover the CNCE gradient.

In both \cref{alg:cd_rnce:rnce_kernel,alg:cd_cnce:cnce_kernel}, a variable $\bernoullivar$ is used to select the next sample in the Markov chain.
However, if we only take a single step of the kernel, as in CD-1, then we can marginalise over $\bernoullivar$ when computing the expected value in \cref{eq:cd:grad}. 
Furthermore, for the CNCE connection, we assume that we average over $\NumNeg$ independent draws from the underlying MCMC kernel.
These measures, which are necessary for an exact equivalence between the two NCE criteria and CD-1, are standard variance reduction techniques for MC estimators.
\section{Insights from CD connection}
\label{sec:extensions}
With the connection between CD and NCE outlined in \cref{prop:cd_rnce:cd_connection,prop:cd_cnce:cd_connection}, we can apply extensions of CD to NCE to improve the performance of the latter.
Apart from the examples described in this section, an obvious extension is that of taking multiple MCMC steps in the kernel, which we leave to \cref{app:nce_extensions}.

\subsection{Choice of proposal distribution $\pNoise$}
\label{sec:adaptive_prop}
For NCE, the proposal distribution $\pNoise$ is an important design choice. In the original interpretation as a proxy-classification problem, an intuitive and common choice is to construct a hard classification problem by choosing $\pNoise$ as similar to the data distribution $\pData$ as possible \citep{cond_nce,ada_nce,flow_CE_of_EBM,how_to_train_ebm}. A choice that has also been theoretically motivated \citep{nce}. 

Our interpretation of both RNCE and CNCE as special cases of CD-1, instead suggests that we should choose $\pNoise$ as close as possible to $\pModel$. Indeed, the proposal distribution is used to construct the kernel meant for estimating $\gradP \log \partFn$ in \cref{eq:unnorm_model:grad}, and this kernel has $\pModel$ as its stationary distribution.
Setting $\pNoise$ as an approximation to $\pModel$ has been proposed before \citep{goodfellow_proposal,ada_nce} and recent work has shown empirically and in some limit cases that $\pData$ is not the optimal proposal distribution \citep{chehab2022optncenoise}.
These results however, apply only to standard NCE and the current literature remains inconclusive, where setting $\pNoise$ close to $\pData$ is still a common choice \citep{nce,cond_nce}.
With the connection between NCE and CD we provide another motivation in favour of the model distribution,
by showing that setting $\pNoise = \pModel$ gives unbiased estimates of the gradient in \cref{eq:unnorm_model:grad}, up to a constant scaling, for both RNCE and CNCE:
\begin{proposition}[Gradient estimate for RNCE with $\pNoise = \pModel$]
\label{thm:q:rank:unbiased}
If $\pNoise = \pModel$, then the expected gradient of the RNCE criterion $\gradP{} \rankLossFn(\param)$ in \cref{eq:rank_nce:grad} is
\begin{align}
  \expect[\pNoise(\yNegSeq{1})]{\gradP{} \rankLossFn(\param)}
  = \frac{J}{\NumNeg + 1} \gradP{} (- \log \pModel(\y_0)).
\end{align}
\end{proposition}

\begin{proposition}[Gradient estimate for CNCE with
$\pNoise = \pModel$]
\label{thm:q:cond:unbiased}
If $q(\cdot \mid \y_0) = \pModel(\cdot)$, independent of $\y_0$, then
the expected gradient of the CNCE criterion $\gradP{} \condLossFn(\param)$ in \cref{eq:cond_nce:grad}
is:
\begin{align}
  &\expect[\pNoise(\yNegSeq{1} \mid \x_0)]{ \gradP{} \condLossFn(\param)} 
   = \frac{1}{2}  \gradP{}(- \log \pModel(\x_0)).
\end{align}
\end{proposition}
See \cref{app:proof:gradient-rnce-cnce} for the proofs.

For RNCE, while \cref{thm:unbiased_cis:log_part_fn} holds for any $\pNoise$, Lemma~\ref{thm:q:rank:unbiased} is stronger in the sense that it indicates that there is an idealised case, i.e., $\pNoise = \pModel$, for which RNCE gives unbiased gradient estimates also when $\y_0 \sim \pData(\cdot)$. In contrast, $\pNoise=\pData$ is not guaranteed to cover the support of $\pModel$, in which case the requirements of \cref{thm:unbiased_cis:log_part_fn} are not fulfilled. 
Of course, neither the data nor model distribution is available for us to evaluate in practice, but it nevertheless gives a guideline for selecting $\pNoise$.

Here we consider a method akin to Markovian Score Climbing \citep{markov_score_climbing} for learning a parameterised proposal $\pNoiseF$ jointly with $\pModel$. With the aim to make $\pNoiseF$ resemble $\pModel$,
we propose to minimise the KL divergence between the two distributions, which is equivalent to minimising the cross-entropy:
\begin{align}
    \label{eq:prop_distr:kl_crit}
    \argMin{\varphi} \klDiv{\pModel}{\pNoiseF} 
    &= \argMin{\varphi} \expect[\pModel(\y')]{ - \log \pNoiseF(\y') } 
    \nonumber \\ & \eqDefRight \argMin{\varphi} \lossFn(\varphi).
\end{align}
Note that we use the divergence from  $\pModel$ to $\pNoiseF$,
since we require $\pNoiseF$ to cover the support of $\pModel$. 
The expectation \wrt $\pModel$ is intractable,
but we already have a method to sample from this distribution: 
$\kernel(\x'|\x_0)$.
For example, we can estimate the gradient with the CIS kernel defined in \cref{alg:cd_rnce:rnce_kernel}:
\begin{align}
    \label{eq:prop_distr:kl_crit:grad}
    \gradF \lossFn(\varphi)
    &\approx \expect[\rankKernel(\y' \mid \y_0)]{ - \gradF \log \pNoiseF(\y') }
    \nonumber \\ & \approx - \sum_{\numNeg=0}^\NumNeg \wNorm{\numNeg} \gradF \log \pNoiseF(\y_\numNeg) 
    \eqDefRight \gradF \lossFnHat(\varphi).
\end{align}
Therefore, the model $\pModel$ and the proposal $\pNoiseF$ can be estimated simultaneously, using samples from the same kernel $\rankKernel$.

As in \cref{thm:unbiased_cis:log_part_fn}, this estimate is unbiased under the idealised assumption that $\y_0 \sim \pModel(\cdot)$.
\begin{proposition}[Unbiased CIS estimate of $\gradF{} \lossFn(\varphi)$]
    \label{thm:unbiased_cis:prop_loss}
    If $\y_0 \sim \pModel(\cdot)$, then the CIS estimator gives an unbiased estimate of the gradient $\expect[\pModel(\y_0), \pNoiseF(\yNegSeq{1})]{\gradF \lossFnHat(\varphi)}  = \gradF \lossFn(\varphi)$.
\end{proposition}
See \cref{app:proof-adaptive-unbiased} for the proof.

Adapting $\pNoiseF$ towards $\pModel$ has been proposed before,
especially in the field of Adaptive IS \citep{adaptive_is}.
It has also been used for NCE; \citet{ada_nce} proposed it, but as a means of achieving $\pNoiseF \approx \pData$, and \citet{learn_prop} motivates $\pNoiseF \approx \pModel$ when estimating $\pModel$ with ML-IS and then also use this proposal for RNCE.
Our connection to CD provides a theoretical argument for why this is a good design choice.
\subsection{Persistent NCE}
Persistent Contrastive Divergence (PCD) is an extension of CD,
with a modified kernel-based sampling method \citep{pcd}. 
Instead of re-initialising the MCMC chain based on a sample $\y_0 \sim \pData(\cdot)$ at every training iteration, PCD initialises the chain at iteration $\epoch$ using the sampled output at the previous iteration, $\epoch-1$. Only at the start is the chain initialised with an actual data sample. The motivation is that this will improve convergence over standard CD, as the samples from the kernel will lie closer to the model distribution. 

For persistent RNCE and CNCE, we update the Markov chain at iteration $\epoch$ by sampling an actual index $\selectInd$ as in \cref{alg:cd_rnce:rnce_kernel}~or~\ref{alg:cd_cnce:cnce_kernel}. At iteration $\epoch$, we estimate the gradient using $\kernel(\y' \mid \yIt{\epoch}_0)$ in place of $\kernel(\y' \mid \y_0)$ in \cref{eq:cd:grad}, where $\yIt{\epoch}_0 \eqDef \yIt{\epoch-1}_\selectInd$ is a sample from the kernel in the previous iteration, $\epoch-1$.
Note that, while we sample $\selectInd$ to update the Markov chain, we still marginalise over this latent variable when evaluating the expectation in \cref{eq:cd:grad}.
Similarly to \citet{pcd}, when training with 
SGD, we keep track of one continuing chain for each training data point in a batch. For CNCE, this translates to running $\NumNeg$ chains per data point in parallel.

\begin{figure*}
\adjustbox{valign=b}{\begin{subfigure}[b]{.35\textwidth}

\fontsize{20}{12}\selectfont
\begin{tikzpicture}[scale=0.39]
\definecolor{darkgray176}{RGB}{176,176,176}
\definecolor{darkorange25512714}{RGB}{255,127,14}
\definecolor{lightgray204}{RGB}{204,204,204}
\definecolor{steelblue31119180}{RGB}{31,119,180}

\begin{loglogaxis}[
legend cell align={left},
legend style={at={(0.03,0.25)},anchor=west, fill opacity=0.8, draw opacity=1, text opacity=1, draw=lightgray204},
tick align=outside,
tick pos=left,
x grid style={darkgray176},
xlabel={Iter.},
xlabel style={yshift=0.1 cm},
xmin=1,
xmax=6500,
xtick style={color=black},
xticklabel style={yshift=0.0 cm},
y grid style={darkgray176},
ylabel={$\klDiv{\pData}{\pModel}$},
ymin=0.001,
ymax=20,
ytick style={color=black}
]
  \addplot[
     color2,
     error bars/.cd,
     y dir=both,
     error bar style={line width=2pt,solid},
  ]
  table[
      x=t,
      y=kl,
      y error plus=upp,
      y error minus=low,
      col sep=comma
  ]
  {content/experiments/figs/adaptive_q_toy_example/20_runs_p_d.txt};
  \addlegendentry{$\pNoise = \pData$};
  \addplot[
     color1,
     error bars/.cd,
     y dir=both,
     error bar style={line width=2pt,solid},
  ]
  table[
      x=t,
      y=kl,
      y error plus=upp,
      y error minus=low,
      col sep=comma
  ]
  {content/experiments/figs/adaptive_q_toy_example/20_runs_p_t.txt};
  \addlegendentry{$\pNoise = \pModel$};
  \addplot[
     color3,
     error bars/.cd,
     y dir=both,
     error bar style={line width=4pt,solid},
  ]
  table[
      x=t,
      y=kl,
      y error plus=upp,
      y error minus=low,
      col sep=comma
  ]
  {content/experiments/figs/adaptive_q_toy_example/20_runs_q_f.txt};
  \addlegendentry{$\pNoise = \pNoiseF$};
\end{loglogaxis}
\end{tikzpicture}
\end{subfigure}}%
\adjustbox{valign=b}{\begin{subfigure}[b]{.55\textwidth}
\input{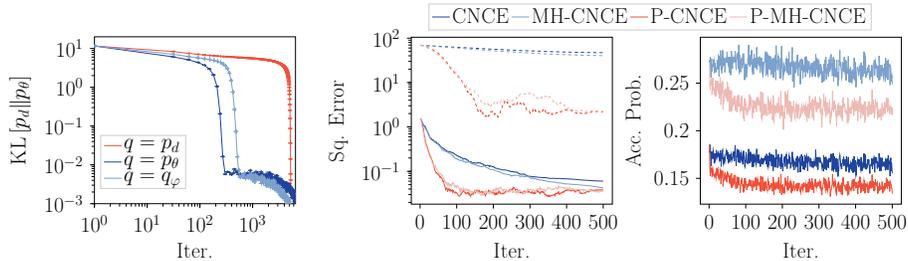}
\end{subfigure}}
\caption{\textbf{Left:} Convergence of $\pModel$ for different choices of proposal distribution $\pNoise$. Here, $\pNoiseF$ is initialised at $\pData$ and we show the median divergence $\klDiv{\pData}{\pModel}$. The error bars mark the 25th and 75th percentile respectively, estimated from 20 repetitions. \textbf{Middle-Right:} Results for ring model experiments reported over training iterations and as median (solid lines) and worst-case (dashed lines) estimated from 100 experiments. Middle: Squared parameter error of CNCE, CNCE with Metropolis--Hastings acceptance probability (MH-CNCE), persistent CNCE (P-CNCE) and persistent MH-CNCE (P-MH-CNCE). Right: Acceptance probability of (P-)CNCE and (P-)MH-CNCE when training with (P-)CNCE.
}
\label{fig:exp:toy_examples}
\end{figure*}
\subsection{MH variant of CNCE}
The kernel used in the CD formulation of the CNCE criterion has similarities with the Metropolis--Hastings (MH) algorithm \citep{mh_metropolis,mh_hastings},
but with a non-standard acceptance probability. In the context of MCMC, and specifically the class of methods proposed by \citet{mh_hastings}, the MH acceptance probability is optimal in terms of Peskun ordering \citep{mh_optimal}. 

With this in mind, we might expect that the MH acceptance probability will improve performance also in CNCE. Hence, we consider CD with the kernel given in \cref{alg:cd_cnce:cnce_kernel}, but with the acceptance probability, $\alpha(\y_0, \y_1) = \wNormCond{1}{0}$, replaced by the standard MH acceptance probability
\begin{align}
    \alpha(\y_0, \y_1) 
    &= \min \left\{1, \wCond{1}{0}/\wCond{0}{1}\right\}.
    \label{eq:mh:acc_prob}
\end{align}
We use a conditional proposal distribution and calculate the weights according to \cref{eq:cnce:weight}.
The kernel will leave $\pModel$ invariant as it fulfils the detailed balance condition \citep{mh_hastings}.
Just as with the original CNCE criterion, we propose to marginalise over the latent variable $\bernoullivar$ and to
use an average over $\NumNeg$ noisy examples to reduce the variance of the Monte Carlo estimate. We refer to this method as MH-CNCE.
\subsection{Sequential Monte Carlo RNCE}
\label{sec:ext:smc_rnce}
Sequential Monte Carlo (SMC) is a generalisation of IS which interleaves IS steps with resampling in a sequential manner; see, \eg, \citep{elements_of_smc}.
SMC is particularly useful for sampling from time series or other sequential models, but can be used more generally \citep{elements_of_smc}.
The interpretation of RNCE as ML-CIS suggests a generalisation of RNCE obtained by replacing CIS with Conditional SMC (CSMC; see \citep{pmcmc}).
Details on this algorithm are given in \cref{app:nce_extensions}.
The resulting SMC-RNCE method has the potential to improve RNCE for sequential models or, wherever SMC is more efficient than IS.
\section{Experiments}
\label{sec:exp:exp}
We provide experiments to empirically test the theoretical results of the paper and to demonstrate the proposed extensions of the NCE criteria.
For additional experiments and details, see \cref{app:exp}
\footnote{Code available at \href{https://github.com/jackonelli/nce_cd_cis/tree/test_merge}{\texttt{github.com/jackonelli/nce\_cd\_cis}} }.

\begin{table*}
  \caption{Results for the autoregressive EBM, given as mean $\pm$ standard error over ten estimates. \textbf{Top:} Test log-likelihood estimated using SMC with $5\cdot 10^6$ samples. \textbf{Bottom:} 2-Wasserstein distance between $\pModel$ and $\pData$, estimated using $1\cdot 10^4$ samples ($2\cdot 10^3$ samples for Miniboone). 
  Samples from $\pModel$ are drawn using SMC.}
  \label{tab:autoregr_res}
  \centering
   \begin{tabular}{l | ccc} 
    & ML-IS & RNCE & SMC-RNCE \\
    \midrule
    Dataset (D) &\multicolumn{3}{c}{Log-likelihood} \\ \midrule
    Power ($6$) & -3.93 $\pm$  0.195 &  0.617 $\pm$ 0.010 & \textbf{0.695 $\pm$ 0.0002} \\
    Gas ($8$) & -2037.2 $\pm$ 0.25 & 2.60 $\pm$ 4.68 & \textbf{13.01 $\pm$ 0.013} \\
    Hepmass ($21$) & -607.42 $\pm$ 48.02 & -14.95 $\pm$ 0.001 & \textbf{-13.47 $\pm$ 0.0004} \\
    Miniboone ($43$) & - & -249.72 $\pm$ 8.70 & \textbf{-15.24 $\pm$ 0.65} \\
    BSDS300 ($63$) & - & \textbf{155.37 $\pm$ 1.09} & 145.73 $\pm$ 1.22 \\ \midrule
    &\multicolumn{3}{c}{Wasserstein distance} \\ \midrule
    Power ($6$) & 255.75 $\pm$  51.54 &  44.11 $\pm$ 1.54 & \textbf{39.32 $\pm$ 1.25} \\
    Gas ($8$) & 5183.4 $\pm$ 4332.5 & 203.69 $\pm$ 12.57 & \textbf{64.02 $\pm$ 8.79} \\
    Hepmass ($21$) & 747.01 $\pm$ 60.25 & 244.84 $\pm$ 0.614 & \textbf{236.70 $\pm$ 0.300} \\
    Miniboone ($43$) & - & 1175.7 $\pm$ 201.52 & \textbf{259.94 $\pm$ 49.74} \\
    BSDS300 ($63$) & - & 85.47 $\pm$ 6.54 & 76.50 $\pm$ 9.34 \\
    \bottomrule
  \end{tabular}
\end{table*}

\subsection{Adaptive proposal distribution}
\label{sec:exp:proposal}
To support the claims of \cref{sec:adaptive_prop}, we conduct a toy experiment where 
$\pData = \normal{\bm{0}, \identityMatrix}$ and
$\pModel = \normal{\bm{\mu}_\param, \Sigma_\param }$
are 5-dimensional multivariate Gaussians, allowing us to sample from and evaluate both distributions exactly.
The model $\pModel$ is parameterised by a mean vector $\bm{\mu}_\param$ and a diagonal covariance matrix $\Sigma_\param$, which are estimated using RNCE.

We study the effect of adapting the proposal distribution to either the data or model distribution.
For reference, we test the idealised cases where $\pNoise = \pData$ and $\pNoise = \pModel$, the former is fixed whereas for the latter we set $\pNoise$ to the current $\pModel$ at every step.
The adaptive proposal $\pNoise = \pNoiseF$ is parameterised the same way as $\pModel$, but with independent parameters $\varphi$. It is jointly estimated with $\pModel$, using the approximation of the gradient in \cref{eq:prop_distr:kl_crit:grad}, which only requires the unnormalised model $\pUnnorm$.
To make the problem more challenging and realistic, we initialise $\pNoiseF$ equal to $\pData$, since we in practice would only have access to samples from $\pData$ for initialisation.

\Cref{fig:exp:toy_examples} shows the convergence of $\pModel$ for the different choices of proposals.
It is clear that $\pNoise = \pModel$ is the best choice.
Interestingly, the (only practically applicable) adaptive proposal $\pNoiseF$ performs much better than using the exact data distribution.
Note that an adaptive proposal which targets the static $\pData$ corresponds to matching the moments of $\pData$ from data, and would therefore be a very close approximation of $\pNoise = \pData$.

\subsection{MH variant and persistent CNCE}
\label{sec:exp:ring_model}
The MH acceptance probability in \cref{eq:mh:acc_prob} is known to perform well in the MCMC setting and we investigate its impact for CNCE, by evaluating the proposed MH-CNCE extension.
We also investigate the performance benefits of applying persistence to CNCE (P-CNCE) as well as to the MH extension (P-MH-CNCE).
To this end, we conduct an experiment similar to the ring model experiment in \citep{cond_nce}.
The unnormalised probability density function (pdf) is given by: $\log \pUnnorm(\y) = -0.5\exp(\param) (\twonorm{\y} - \mu)^2$
with $\quad \y \in \mathbb{R}^5$ and $\twonorm{\cdot}$ the Euclidean norm.
We seek to estimate the log precision $\param$,
while the mean $\mu$ is known.

We train models using SGD, with $N=200$ data samples drawn from the true pdf and $\NumNeg=5$ samples from $\pNoise$. The proposal distribution, $\pNoise$, is a Gaussian, centred at the data sample, $\y_0$, and with a diagonal covariance matrix, $\pNoise(\y_1 \mid \y_0) = \normal{\y_1; \y_0, \epsilon^2 \identityMatrix}$. The parameter $\epsilon$ is a hyperparameter, estimated as the mean standard deviation of the training data. 

During training, we measure the squared error of the estimated precision, $\exp(\param)$. \Cref{fig:exp:toy_examples} shows the median and worst-case squared error obtained over 100 experiments. To assess the difference between the CNCE and MH-CNCE acceptance probabilities, we additionally track these quantities and report their median in \cref{fig:exp:toy_examples}. For the comparison to be reasonable, we need to evaluate the probabilities on the same set of samples and hence show the acceptance probabilities obtained when training with CNCE (or P-CNCE for PCD), but where we also calculate the MH acceptance probability. The trends are similar when training with (P-)MH-CNCE and evaluating both probabilities.

The MH acceptance probability is indeed larger than the one used by CNCE, as confirmed by \cref{fig:exp:toy_examples}.
This also seems to lead to slightly faster converge, at least for standard CNCE. Even when the improvement in performance is small, changing the acceptance probability comes without any additional costs. Further improvements over both CNCE and MH-CNCE are seen by adding persistence, in terms of convergence rate as well as both median and worst-case performance.
\subsection{Autoregressive EBM}\label{sec:arebm}

Inspired by \cite{aem,ace}, we perform experiments with an autoregressive EBM (AR-EBM). Specifically, we factorise the model as
\(
    p_\theta(\y) = \frac{1}{\partFn}\prod_{\xdim=1}^{\Xdim} \pUnnorm(\y_{\xdim}\mid \y_{1:(\xdim-1)}),  
\)
for a given ordering of the $\Xdim$ features $\y_{\xdim}$, $\xdim=1, \ldots, \Xdim$, and where $\y_{1:(\xdim-1)} = [\y_{1}, \ldots, \y_{\xdim-1}]^\top$.
The AR-EBM predicts the energy $-\log \pUnnorm(\y_{\xdim} \mid  \y_{1:(\xdim-1)})$ of feature $\xdim$ conditional on the preceding features $\y_{1:(\xdim-1)}$. 

We learn a parameterised proposal distribution $\pNoise_\varphi$, together with the AR-EBM.
The proposal distribution also has an autoregressive factorisation and each factor is a Gaussian Mixture Model (GMM) with 10 components.
In the experiments, both the AR-EBM and the proposal are parameterised by fully-connected neural networks with residual connections \citep{res_mapping}, see \cref{app:exp} for details.

The AR-EBM has a sequential structure that can be leveraged by an SMC algorithm. We therefore 
compare the following methods for training the model: ML-IS, RNCE, and the proposed SMC extension of RNCE (SMC-RNCE, \cref{sec:ext:smc_rnce}).
We perform experiments on four datasets (Power, Gas, Hepmass and Miniboone) from the UCI machine learning repository \citep{uci} as well as the BSDS300 dataset \citep{BSDS300}, pre-processed according to \citet{aem}.

All methods use a total of $\NumNeg=20$ samples from the proposal, either to estimate the log-normalisation constant in \cref{eq:unnorm_model:ll} (ML-IS) or as negative examples (RNCE/SMC-RNCE).
In \cref{tab:autoregr_res}, we report test log-likelihoods, estimated using SMC, as well as estimated 2-Wasserstein distances \citep{villani2009opttransp} between $\pModel$ and $\pData$ for each AR-EBM.

Results for ML-IS are omitted for the two datasets of highest dimension, (Miniboone and BSDS300), 
as we found training to be highly unstable. As expected, we observe a performance advantage of the proposed SMC-RNCE criterion, and particularly for the Gas and Miniboone datasets.
We also observe an advantage of RNCE over ML-IS, as suggested by the established equivalence between RNCE and ML-CIS.
\section{Conclusion}
In this paper, we contributed to building a more coherent framework for the estimation of unnormalised models, by linking the proxy-criterion noise-contrastive estimation (NCE) to approximate maximum likelihood (ML) methods. Firstly, we established that ranking NCE is equivalent to ML estimation with conditional importance sampling (CIS). This equivalence gives a possible explanation for why ranking NCE would perform better than ML estimation with standard importance sampling; the gradient of the ranking NCE criterion is an approximately unbiased gradient of the log-likelihood. Secondly, we derived ranking NCE and conditional NCE as special cases of contrastive divergence, using MCMC kernels based on CIS and a Metropolis–Hastings-like method, respectively. The established links provide theoretical support for why the optimal noise distribution in NCE is the model, and not the data, distribution and we propose a practical method for adapting the proposal to this distribution. 
Moreover, our integration of NCE into a standard MCMC setting enables the use of more robust MCMC approaches while at the same time preserving the efficiency and simplicity of NCE. We propose several extensions to the NCE methods and showcase their potential to improve model performance.
\subsubsection*{Acknowledgments}
This research is financially supported by the Swedish Research Council via the project
\emph{Handling Uncertainty in Machine Learning Systems} (contract number: 2020-04122),
the Wallenberg AI, Autonomous Systems and Software Program (WASP) funded by the Knut and Alice Wallenberg Foundation,
and
the Excellence Center at Linköping--Lund in Information Technology (ELLIIT).

\bibliography{refs.bib}

\begin{thebibliography}{41}
\providecommand{\natexlab}[1]{#1}
\providecommand{\url}[1]{\texttt{#1}}
\expandafter\ifx\csname urlstyle\endcsname\relax
  \providecommand{\doi}[1]{doi: #1}\else
  \providecommand{\doi}{doi: \begingroup \urlstyle{rm}\Url}\fi

\bibitem[Andrieu et~al.(2010)Andrieu, Doucet, and Holenstein]{pmcmc}
C.~Andrieu, A.~Doucet, and R.~Holenstein.
\newblock Particle {M}arkov chain {M}onte {C}arlo methods.
\newblock \emph{Journal of the Royal Statistical Society Series B}, 72\penalty0 (3):\penalty0 269--342, 2010.

\bibitem[Asuncion et~al.(2010)Asuncion, Liu, Ihler, and Smyth]{mcmc_mle}
A.~U. Asuncion, Q.~Liu, A.~T. Ihler, and P.~Smyth.
\newblock Particle filtered {MCMC-MLE} with connections to contrastive divergence.
\newblock In \emph{{{{I}nternational {C}onference on {M}achine {L}earning}}}, 2010.

\bibitem[Barker(1965)]{barkers_method}
A.~A. Barker.
\newblock {{M}onte {C}arlo calculations of the radial distribution functions for a proton-electron plasma}.
\newblock \emph{{Australian Journal of Physics}}, 18\penalty0 (2):\penalty0 119--133, 1965.

\bibitem[Bottou et~al.(2018)Bottou, Curtis, and Nocedal]{optim_large_scale}
L.~Bottou, F.~E. Curtis, and J.~Nocedal.
\newblock Optimization methods for large-scale machine learning.
\newblock \emph{{SIAM} Review}, 60\penalty0 (2):\penalty0 223--311, 2018.

\bibitem[Bugallo et~al.(2017)Bugallo, Elvira, Martino, Luengo, Miguez, and Djuric]{adaptive_is}
M.~F. Bugallo, V.~Elvira, L.~Martino, D.~Luengo, J.~Miguez, and P.~M. Djuric.
\newblock Adaptive importance sampling: The past, the present, and the future.
\newblock \emph{IEEE Signal Processing Magazine}, 34\penalty0 (4):\penalty0 60--79, 2017.

\bibitem[Carreira-Perpinan and Hinton(2005)]{on_cd_learning}
M.~A. Carreira-Perpinan and G.~Hinton.
\newblock On contrastive divergence learning.
\newblock In \emph{International Workshop on Artificial Intelligence and Statistics}, 2005.

\bibitem[Ceylan and Gutmann(2018)]{cond_nce}
C.~Ceylan and M.~U. Gutmann.
\newblock Conditional noise-contrastive estimation of unnormalised models.
\newblock In \emph{International Conference on Machine Learning}, 2018.

\bibitem[Chehab et~al.(2022)Chehab, Gramfort, and Hyvarinen]{chehab2022optncenoise}
O.~Chehab, A.~Gramfort, and A.~Hyvarinen.
\newblock The optimal noise in noise-contrastive learning is not what you think.
\newblock In \emph{Conference on Uncertainty in Artificial Intelligence}, 2022.

\bibitem[Du et~al.(2021)Du, Li, Tenenbaum, and Mordatch]{improved_cd}
Y.~Du, S.~Li, J.~Tenenbaum, and I.~Mordatch.
\newblock {Improved Contrastive Divergence Training of Energy Based Models}.
\newblock In \emph{International Conference on Machine Learning}, 2021.

\bibitem[Florence et~al.(2022)Florence, Lynch, Zeng, Ramirez, Wahid, Downs, Wong, Lee, Mordatch, and Tompson]{implicit_cloning}
P.~Florence, C.~Lynch, A.~Zeng, O.~A. Ramirez, A.~Wahid, L.~Downs, A.~Wong, J.~Lee, I.~Mordatch, and J.~Tompson.
\newblock Implicit behavioral cloning.
\newblock In \emph{Conference on Robot Learning}, 2022.

\bibitem[Gao et~al.(2020)Gao, Nijkamp, Kingma, Xu, Dai, and Wu]{flow_CE_of_EBM}
R.~Gao, E.~Nijkamp, D.~P. Kingma, Z.~Xu, A.~M. Dai, and Y.~N. Wu.
\newblock Flow contrastive estimation of energy-based models.
\newblock In \emph{IEEE/CVF Conference on Computer Vision and Pattern Recognition}, 2020.

\bibitem[Goodfellow(2015)]{goodfellow_proposal}
I.~Goodfellow.
\newblock On distinguishability criteria for estimating generative models.
\newblock In \emph{Workshop Contribution in International Conference on Learning Representations}, 2015.

\bibitem[{Gustafsson} et~al.(2020){Gustafsson}, {Danelljan}, {Timofte}, and {Sch{\"o}n}]{how_to_train_ebm}
F.~K. {Gustafsson}, M.~{Danelljan}, R.~{Timofte}, and T.~B. {Sch{\"o}n}.
\newblock How to train your energy-based model for regression.
\newblock In \emph{British Machine Vision Virtual Conference}, 2020.

\bibitem[Gustafsson et~al.(2022)Gustafsson, Danelljan, and Sch\"{o}n]{learn_prop}
F.~K. Gustafsson, M.~Danelljan, and T.~B. Sch\"{o}n.
\newblock Learning proposals for practical energy-based regression.
\newblock In \emph{International Conference on Artificial Intelligence and Statistics}, pages 4685--4704, 2022.

\bibitem[Gutmann and Hirayama(2011)]{bregman_div_est_unnorm}
M.~Gutmann and J.~Hirayama.
\newblock Bregman divergence as general framework to estimate unnormalized statistical models.
\newblock In \emph{Conference on Uncertainty in Artificial Intelligence}, 2011.

\bibitem[Gutmann and Hyv{\"a}rinen(2012)]{nce}
M.~Gutmann and A.~Hyv{\"a}rinen.
\newblock Noise-contrastive estimation of unnormalized statistical models, with applications to natural image statistics.
\newblock \emph{Journal of Machine Learning Research}, 13:\penalty0 307--361, 2012.

\bibitem[Hastings(1970)]{mh_hastings}
W.~K. Hastings.
\newblock {M}onte {C}arlo sampling methods using {M}arkov chains and their applications.
\newblock \emph{Biometrika}, 57\penalty0 (1):\penalty0 97--109, 1970.

\bibitem[He et~al.(2016)He, Zhang, Ren, and Sun]{res_mapping}
K.~He, X.~Zhang, S.~Ren, and J.~Sun.
\newblock Identity mappings in deep residual networks.
\newblock In \emph{European Conference on Computer Vision}, 2016.

\bibitem[Hinton(2002)]{contr_div_hinton}
G.~E. Hinton.
\newblock Training products of experts by minimizing contrastive divergence.
\newblock \emph{Neural Comput.}, 14\penalty0 (8):\penalty0 1771–1800, Aug 2002.

\bibitem[Hyv{\"a}rinen(2007)]{connection_cd_sm}
A.~Hyv{\"a}rinen.
\newblock Connections between score matching, contrastive divergence, and pseudolikelihood for continuous-valued variables.
\newblock \emph{IEEE Transactions on Neural Networks}, 18\penalty0 (5):\penalty0 1529--1531, 2007.

\bibitem[Jozefowicz et~al.(2016)Jozefowicz, Vinyals, Schuster, Shazeer, and Wu]{nce_ranking_obj}
R.~Jozefowicz, O.~Vinyals, M.~Schuster, N.~Shazeer, and Y.~Wu.
\newblock Exploring the limits of language modeling.
\newblock \emph{arXiv preprint arXiv:1602.02410}, 2016.

\bibitem[Kelly et~al.()Kelly, Longjohn, and Nottingham]{uci}
M.~Kelly, R.~Longjohn, and K.~Nottingham.
\newblock {UCI} machine learning repository.
\newblock URL \url{http://archive.ics.uci.edu}.

\bibitem[{K}ingma and {B}a {L}ei(2015)]{adam}
D.~P. {K}ingma and J.~{B}a {L}ei.
\newblock {A}dam: {A} method for stochastic optimization.
\newblock In \emph{{International Conference on Learning Representations}}, 2015.

\bibitem[K\"{o}ster et~al.(2009)K\"{o}ster, Lindgren, and Hyv\"{a}rinen]{mrfs_score_matching}
U.~K\"{o}ster, J.~T. Lindgren, and A.~Hyv\"{a}rinen.
\newblock Estimating {M}arkov random field potentials for natural images.
\newblock In \emph{International Conference on Independent Component Analysis and Signal Separation}. Springer-Verlag, 2009.

\bibitem[LeCun et~al.(2006)LeCun, Chopra, Hadsell, Ranzato, Huang, and et~al.]{ebm_og}
Y.~LeCun, S.~Chopra, R.~Hadsell, M.~Ranzato, F.~Huang, and et~al.
\newblock A tutorial on energy-based learning.
\newblock \emph{Predicting structured data}, 2006.

\bibitem[Ma and Collins(2018)]{nce_neg_sampl_cond_models}
Z.~Ma and M.~Collins.
\newblock Noise contrastive estimation and negative sampling for conditional models: Consistency and statistical efficiency.
\newblock In \emph{Conference on Empirical Methods in Natural Language Processing}, 2018.

\bibitem[Martin et~al.(2001)Martin, Fowlkes, Tal, and Malik]{BSDS300}
D.~Martin, C.~Fowlkes, D.~Tal, and J.~Malik.
\newblock A database of human segmented natural images and its application to evaluating segmentation algorithms and measuring ecological statistics.
\newblock In \emph{IEEE International Conference on Computer Vision}, 2001.

\bibitem[Metropolis et~al.(1953)Metropolis, Rosenbluth, Rosenbluth, and Teller]{mh_metropolis}
N.~Metropolis, A.~W. Rosenbluth, M.~N. Rosenbluth, and A.~H. Teller.
\newblock Equation of state calculations by fast computing machines.
\newblock \emph{The Journal of Chemical Physics}, 21\penalty0 (6):\penalty0 1087--1092, 1953.

\bibitem[Naesseth et~al.(2020)Naesseth, Lindsten, and Blei]{markov_score_climbing}
C.~Naesseth, F.~Lindsten, and D.~Blei.
\newblock Markovian score climbing: Variational inference with $\text{KL}(p\vert \vert q)$.
\newblock In \emph{Advances in Neural Information Processing Systems}, 2020.

\bibitem[Naesseth et~al.(2019)Naesseth, Lindsten, and Sch\"{o}n]{elements_of_smc}
C.~A. Naesseth, F.~Lindsten, and T.~B. Sch\"{o}n.
\newblock Elements of sequential {M}onte {C}arlo.
\newblock \emph{Foundations and Trends in Machine Learning}, 12\penalty0 (3):\penalty0 307–392, Nov 2019.

\bibitem[Nash and Durkan(2019)]{aem}
C.~Nash and C.~Durkan.
\newblock Autoregressive energy machines.
\newblock In \emph{International Conference on Machine Learning}, 2019.

\bibitem[Peskun(1973)]{mh_optimal}
P.~Peskun.
\newblock Optimum {M}onte-{C}arlo sampling using {M}arkov chains.
\newblock \emph{Biometrika}, 60\penalty0 (3):\penalty0 607--612, 1973.

\bibitem[Pihlaja et~al.(2010)Pihlaja, Gutmann, and Hyv\"{a}rinen]{nce_follow_up}
M.~Pihlaja, M.~Gutmann, and A.~Hyv\"{a}rinen.
\newblock A family of computationally efficient and simple estimators for unnormalized statistical models.
\newblock In \emph{Conference on Uncertainty in Artificial Intelligence}, page 442–449. AUAI Press, 2010.

\bibitem[Robert et~al.(1999)Robert, Casella, and Casella]{mc_stat_met}
C.~P. Robert, G.~Casella, and G.~Casella.
\newblock \emph{{{M}onte {C}arlo statistical methods}}.
\newblock Springer, 1999.

\bibitem[Strauss and Oliva(2021)]{ace}
R.~Strauss and J.~B. Oliva.
\newblock Arbitrary conditional distributions with energy.
\newblock \emph{Advances in Neural Information Processing Systems}, 34, 2021.

\bibitem[Tieleman(2008)]{pcd}
T.~Tieleman.
\newblock Training restricted boltzmann machines using approximations to the likelihood gradient.
\newblock In \emph{{{I}nternational {C}onference on {M}achine {L}earning}}, 2008.

\bibitem[Uehara et~al.(2020)Uehara, Kanamori, Takenouchi, and Matsuda]{unified_est_framework_unnorm_models}
M.~Uehara, T.~Kanamori, T.~Takenouchi, and T.~Matsuda.
\newblock A unified statistically efficient estimation framework for unnormalized models.
\newblock In \emph{International Conference on Artificial Intelligence and Statistics}, 2020.

\bibitem[Villani(2009)]{villani2009opttransp}
C.~Villani.
\newblock \emph{The Wasserstein distances}, pages 93--111.
\newblock Springer Berlin Heidelberg, Berlin, Heidelberg, 2009.
\newblock ISBN 978-3-540-71050-9.
\newblock \doi{10.1007/978-3-540-71050-9_6}.
\newblock URL \url{https://doi.org/10.1007/978-3-540-71050-9_6}.

\bibitem[Welling et~al.(2003)Welling, Mnih, and Hinton]{cd_wormholes}
M.~Welling, A.~Mnih, and G.~E. Hinton.
\newblock Wormholes improve contrastive divergence.
\newblock \emph{{Advances in Neural Information Processing Systems}}, 2003.

\bibitem[Xu(2022)]{ada_nce}
N.~Xu.
\newblock Self-adapting noise-contrastive estimation for energy-based models.
\newblock Master's thesis, Tsinghua University, 2022.

\bibitem[Yair and Michaeli(2021)]{cd_time_rev}
O.~Yair and T.~Michaeli.
\newblock Contrastive divergence learning is a time reversal adversarial game.
\newblock In \emph{International Conference on Learning Representations}, 2021.

\end{thebibliography}
\bibliographystyle{abbrvnat} 

\clearpage
\newpage
\appendix

\section{Theoretical derivations}
\subsection{Gradient of the negative log-likelihood with the IS estimator} \label{app:ml-is-gradient}
In maximum likelihood estimation with importance sampling (ML-IS),
we approximate the normalisation constant $\partFn$ in \cref{eq:unnorm_model:ll} with the IS estimator,
defined in \cref{eq:is:z_hat}, using $\NumNeg$ samples $\y_\numNeg \sim \pNoise(\cdot)$, $\numNeg=1, \ldots, \NumNeg$.
The gradient of this approximation is
\begin{align*}
  -\gradP{}  \log \pModel(\y_0) 
  &\approx - \gradP{} \log \pUnnorm(\y_0) + \gradP{} \log \partFnHatIs.
\end{align*}
Using the definition of the weights $\w(\y_{\numNeg})$ in \cref{eq:is:z_hat},
we write the gradient of the estimated log-normalisation constant:
\begin{align}
  \label{eq:is:estimator:grad:derivation}
  \gradP{} \log \partFnHatIs
  &=\frac{1}{ \partFnHatIs } \gradP{} \partFnHatIs
  =\frac{1}{ \NumNeg \partFnHatIs } \sum_{\numNeg=1}^{\NumNeg} \gradP{} \w(\y_{\numNeg})
  =\frac{1}{ \NumNeg \partFnHatIs } \sum_{\numNeg=1}^{\NumNeg} \w(\y_{\numNeg}) \gradP{} \log \w(\y_{\numNeg}) \nonumber \\
  &=\frac{1}{ \NumNeg \partFnHatIs } \sum_{\numNeg=1}^{\NumNeg} \w(\y_{\numNeg}) \gradP{} \log \pUnnorm(\y_{\numNeg})
  =\sum_{\numNeg=1}^{\NumNeg} \frac{\w(\y_{\numNeg})}{ \sum_{\ell=1}^{\NumNeg} \w(\y_\ell) }
    \gradP{} \log \pUnnorm(\y_{\numNeg}).
\end{align}
Note that this gradient is a self-normalised estimate of the desired gradient, $\mathbb{E}_{\pModel(\y)} [\gradP \log \pUnnorm(\y)]$, and will therefore typically be biased \citep{mc_stat_met}.
Note also that the normalisation is only done over the samples from $\pNoise(\cdot)$ and differs from the normalised weight $\wNorm{\numNeg}$ as defined in \cref{eq:rank_nce:grad}, where the sum is over $\numNeg=0,\,\dots,\,\NumNeg$.

\subsection{RNCE criterion derivation}
\label{app:rnce-criterion}
The RNCE criterion by \citet{nce_ranking_obj} is based on a multi-class classification problem with a single true data point and multiple noisy ones.
Recall, we have $\y_0 \sim \pData(\cdot)$ and noisy samples $\y_j \sim \pNoise(\cdot), \numNeg = 1, \dots, \NumNeg$.
We prepend $\y_0$ to the noisy samples and define $\yNegSeq{0} = [\y_0, \y_1, \dots, \y_{\NumNeg}]$. 

Assume we forget the origin of $\y_0, \yNegSeq{1}$. Let the variable $z \in \{0, \ldots, \NumNeg\}$ denote the index, or class, of the true data sample, and assume that all outcomes are equally probable a priori, i.e., $p(z=\numNeg)=1/(\NumNeg+1)$ for $\numNeg = 0,1,\dots, \NumNeg$. Conditioned on $\yNegSeq{0}$, we want the model to maximise the posterior probability of $z = 0$:
\begin{align}
  \label{eq:nce:categorical}
  \p(z = 0 \mid \yNegSeq{0}) 
  &= \frac{\p(\yNegSeq{1} \mid z = 0) p( \y_0 \mid z = 0) p(z = 0) }{ p(\yNegSeq{0}) } \nonumber \\
  &= \frac{\pModel(\y_0) \prod_{j=1}^{\NumNeg} \pNoise(\y_{\numNeg}) p(z=0) }{ \sum_{\numNeg=0}^{\NumNeg} \pModel(\y_{\numNeg}) \prod_{\ell \neq \numNeg} \pNoise(\y_{\ell}) p(z=j)} \nonumber 
  \\
  &= \frac{\pModel(\y_0) \prod_{\numNeg=1}^{\NumNeg} \pNoise(\y_{\numNeg}) }{ \sum_{\numNeg=0}^{\NumNeg} \pModel(\y_{\numNeg}) \prod_{\ell \neq \numNeg} \pNoise(\y_{\ell}) } \nonumber \\
  &= \left\{ \text{Divide num. and den. by } 
  \frac{1}{\partFn}
  \prod_{\numNeg=0}^{\NumNeg} \pNoise(\y_{\numNeg}) \right\} \nonumber \\
  &=\frac{ \pUnnorm(\y_0) / \pNoise(\y_{0}) }{ \sum_{\numNeg=0}^{\NumNeg} \pUnnorm(\y_\numNeg) / \pNoise(\y_{\numNeg}) } 
  =\frac{ \w(\y_{0}) }{ \sum_{\numNeg=0}^{\NumNeg}  \w(\y_{\numNeg})}.
\end{align}
The RNCE criterion in \cref{eq:rank_nce:crit} follows from minimising the negative logarithm of this probability.
\subsection{Proof of \cref{thm:rank_nce:cis}}
\label{app:proof-rnce-cis}
First, we derive an expression for the gradient of the RNCE criterion in terms of the CIS estimator $\partFnHatCis$ in \cref{eq:cis:z_hat}. We use the expression for the RNCE criterion in \cref{eq:rank_nce:crit}:
%
\begin{align}
  \label{eq:rank_nce:grad:derivation:first}
  \gradP{} \rankLossFn(\param, \y_{0:\NumNeg}) 
  &= \gradP{} \left( - \log \w(\y_0) + \log \sum_{\numNeg=0}^\NumNeg \w(\y_\numNeg) \right)  \nonumber \\
  &= \gradP{} \left( - \log \w(\y_0) + \log (\NumNeg+1) \partFnHatCis \right) \nonumber \\
  &=- \gradP{} \log \pUnnorm(\y_0) + \gradP{} \log \pNoise(\y_0) + \gradP{} \log \partFnHatCis + \gradP{} \log (\NumNeg+1) \nonumber \\
  &=- \gradP{} \log \pUnnorm(\y_0) + \gradP{} \log \partFnHatCis.
\end{align}

This is equivalent to the gradient of the negative log-likelihood in \cref{eq:unnorm_model:ll}, with the normalisation constant estimated with CIS: 
\begin{align}
  \gradP{} (- \log \pModel(\y_0))
  &\approx \gradP{}( - \log \pUnnorm(\y_0) + \log \partFnHatCis) \nonumber \\
  &= - \gradP{} \log \pUnnorm(\y_0) + \gradP{} \log \partFnHatCis.
\end{align}
This concludes the proof. However, for completeness, we also demonstrate that the expression in \cref{eq:rank_nce:grad:derivation:first} matches the expression in \cref{eq:rank_nce:grad}. The second term in the last equality of  \cref{eq:rank_nce:grad:derivation:first} can be expressed as
\begin{align}
  \label{eq:cis:part_fn:grad_log}
  \gradP{} \log \partFnHatCis
  &= \frac{1}{\partFnHatCis} \gradP{} \partFnHatCis
  =\frac{1}{(\NumNeg+1) \partFnHatCis} \sum_{\numNeg=0}^{\NumNeg} \gradP{} \w(\y_\numNeg) \nonumber \\
  &=\frac{1}{(\NumNeg+1) \partFnHatCis} \sum_{\numNeg=0}^{\NumNeg} \w(\y_\numNeg) \gradP{} \log \w(\y_\numNeg)  \nonumber \\
  &=\sum_{\numNeg=0}^{\NumNeg} \frac{\w(\y_\numNeg) }{\sum_{\ell=0}^{\NumNeg}\w(\y_\ell) } \gradP{} \log \w(\y_\numNeg) \nonumber \\
  &=\sum_{\numNeg=0}^{\NumNeg} \wNorm{\numNeg} \gradP{} \log \w(\y_\numNeg)
  =\sum_{\numNeg=0}^{\NumNeg} \wNorm{\numNeg} \gradP{} \log \pUnnorm(\y_\numNeg).
\end{align}
For the last equality, we use $\gradP{} \log \pNoise(\x_{\numNeg}) =0$, assuming that $\pNoise$ is independent of $\theta$. Clearly, plugging \cref{eq:cis:part_fn:grad_log} into \cref{eq:rank_nce:grad:derivation:first} yields the expression in \cref{eq:rank_nce:grad}.
\subsection{Proof of \cref{thm:unbiased_cis:log_part_fn}}
\label{app:proof:prop-unbiased-cis}
To compute the CIS estimator of the normalisation constant $\partFnHatCis$ we first sample an index $z$ uniformly, such that
\begin{align}
    p(z = i) = \begin{cases}
\frac{1}{\NumNeg+1} & \text{if } i \in \{0,1,\dots,\NumNeg \}, \\
0 & \text{otherwise},
    \end{cases}
\end{align}
and then sample 
\begin{align}
    \y_{z} &\sim \pModel(\y), \nonumber \\ 
    \y_\numNeg &\sim \pNoise(\y),\, \numNeg \neq z, \numNeg = 0, \dots, \NumNeg. 
\end{align}
To simplify the notation, we introduce $\y_{-z} \eqDef [\y_0, \dots, \y_{z-1}, \y_{z+1}, \dots, \y_\NumNeg]$. Note that we in the main paper, without loss of generalisation, fix $z=0$ (we can just re-order the indices).

First, we prove a general property of the CIS estimator:
\begin{lemma}[Unbiased general CIS estimate]
    \label{thm:unbiased_cis:general} 
    Assume $\y_{z} \sim \pModel(\cdot)$ and $\y_{-z} \sim \pNoise(\cdot)$.
    Then, for any function $f$ and deterministic index $i \in \{0,\, \dots,\, \NumNeg\}$
    \begin{align}
        \expect[p(z, \yNegSeq{0})]{\frac{f(\y_i)}{\partFnHatCis}} = \frac{1}{\partFn}\expect[\pNoise(\y)]{f(\y)}.
    \end{align}
\end{lemma}
\begin{proof}
We can write the joint distribution as
\begin{align}
  \label{eq:cis:joint_pdf}
  \p(z, \yNegSeq{0})
  &= \overbrace{\frac{1}{\NumNeg + 1 }}^{\p(z)}
  \times \overbrace{\pModel(\y_{z})}^{\p(\y_{z} \vert z)}
  \times \overbrace{ \prod_{\numNeg \neq z} \pNoise(\y_{\numNeg}) }^{ \p(\ySeqNot{z}) }
\end{align}
Then, we find the marginal distribution of $\yNegSeq{0}$ as
\begin{align}
     \p(\yNegSeq{0})
  &= \sum_{z=0}^{\NumNeg}\frac{1}{\NumNeg + 1 }
  \pModel(\y_{z})
  \prod_{\numNeg \neq z} \pNoise(\y_{\numNeg})
  = \sum_{z=0}^{\NumNeg} \frac{1}{(\NumNeg + 1) \partFn} \w(\y_{z}) \prod_{\numNeg = 0}^{\NumNeg} \pNoise(\y_\numNeg) \nonumber \\
  &= \frac{1}{(\NumNeg + 1) \partFn}  \prod_{\numNeg = 0}^{\NumNeg} \pNoise(\y_\numNeg) \sum_{z=0}^{\NumNeg} \w(\y_{z})
  = \frac{\partFnHatCis }{\partFn}  \prod_{\numNeg = 0}^{\NumNeg} \pNoise(\y_\numNeg) 
\end{align}
and from \cref{eq:cis:joint_pdf}
\begin{align}
    p(z\mid \y_{0:\NumNeg}) = \frac{p(z, \y_{0:\NumNeg})}{p(\y_{0:\NumNeg})} = \frac{\w(\y_{z})}{\sum_{\ell=0}^{\NumNeg} \w(\y_{\ell})}.
\end{align}

Then, for any function $f(\cdot)$ and deterministic index $i$ we have:
\begin{align}
    \label{eq:cis:expectation:general}
    \expect[\p(z, \yNegSeq{0})]{ \frac{f(\y_{i})}{\partFnHatCis} }     
    & = \expect[\p(\yNegSeq{0})]{ \frac{f(\y_{i})}{\partFnHatCis}}\nonumber
    = \int \frac{f(\y_{i})}{\partFnHatCis} \frac{\partFnHatCis}{\partFn} \prod_{\numNeg = 0}^{\NumNeg} \pNoise(\y_{\numNeg}) d\y_{0:\NumNeg} \nonumber \\
    & = \frac{1}{\partFn}\int f(\y_{i}) \pNoise(\y_{i})d\y_i \prod_{\numNeg \neq i} \pNoise(\y_{\numNeg}) d\y_{-i}  \nonumber \\
    &= \frac{1}{\partFn} \expect[\pNoise]{f(\y)} \int \prod_{\numNeg \neq i} \pNoise(\y_{\numNeg}) d\y_{-i} \nonumber \\
    &= \frac{1}{\partFn} \expect[\pNoise]{f(\y)},
\end{align}
where the first equality follows from the fact that the integrand does not depend on $z$.
\end{proof}

To prove \cref{thm:unbiased_cis:log_part_fn}, we first note that
\begin{align}
  \label{eq:cd_cis:grad_estimation}
  \expect[\pModel(\y_0) \pNoise(\yNegSeq{1})]{\gradP{} \log \partFnHatCis }
  &=\expect[\p(z, \yNegSeq{0})]{\gradP{} \log \partFnHatCis }  \nonumber \\
  &=\expect[\p(z, \yNegSeq{0})]{\frac{1}{\partFnHatCis} \gradP{} \partFnHatCis }
  \nonumber \\
  &= \frac{1}{\NumNeg + 1} \sum_{\numNeg = 0}^{\NumNeg} \expect[\p(z, \yNegSeq{0})]{\frac{\gradP{ \w(\y_\numNeg) } }{\partFnHatCis} }.
\end{align}
From \cref{thm:unbiased_cis:general} we have
\begin{align}
  \expect[\p(z, \yNegSeq{0})]{\frac{\gradP{ \w(\y_\numNeg) } }{\partFnHatCis} }
  &= \frac{1}{\partFn} \expect[\pNoise]{ \gradP{ \w(\y) } } \nonumber \\
  &=\frac{1}{\partFn} \int \w(\y) \left[ \gradP{ \log \w(\y) } \right] \pNoise(\y) \d \y \nonumber \\
  &=\frac{1}{\partFn} \int \frac{\partFn \pModel(\y)}{\pNoise(\y)} \left[ \gradP{ \log \frac{\partFn \pModel(\y)}{\pNoise(\y)} } \right]  \pNoise(\y) \d \y \nonumber \\
  &=\int \pModel(\y) \left[ \gradP{ \log \partFn } + \gradP{ \log \pModel(\y) } - \gradP{ \log \pNoise(\y) } \right] \d \y \nonumber \\
  &= \gradP{ \log \partFn }. 
\end{align}
The middle term in the second to last row is zero since:
\begin{align}
    \int \pModel(\y) \gradP{ \log \pModel(\y) } \d \y
    &=\int \gradP{ \pModel(\y) } \d \y
    =\gradP{} \int \pModel(\y) \d \y
    =\gradP{} 1
    = 0,
\end{align}
and $\gradP \log \pNoise(\y)=0$ if $\pNoise(\y)$ is independent of $\param$.

Finally, we conclude the proof with
\begin{align}
  \expect[\pModel(\y_0) \pNoise(\yNegSeq{1})]{\gradP{} \log \partFnHatCis }
  &= \frac{1}{\NumNeg + 1} \sum_{\numNeg = 0}^\NumNeg \gradP{} \log \partFn
  = \gradP{} \log \partFn.
\end{align}
That is, we get an unbiased estimate of the $\gradP{} \log \partFn$ if $\pNoise$ is independent of $\param$ and if the estimate is computed using a sample from the distribution $\p(z, \yNegSeq{0})$ defined above, corresponding to the CIS procedure where we sample from $\pModel(\y_0)\pNoise(\yNegSeq{1})$.
\subsection{Proof of \cref{prop:cd_rnce:cd_connection}}
\label{app:proof-cd-rnce}
The expectation \wrt $\rankKernel$ in \cref{eq:cd:grad}, starting at a data sample $\y_0 \sim \pData$, is given by
\begin{align}
    \textstyle
    \label{eq:rnce:cd_derivation}
    \expect[\rankKernel(\y' \vert \y_0)]{\gradP \log \pUnnorm(\y') }
    &= \expect[\pNoise(\y_{1:\NumNeg})]{ \expect[\categorical{\selectInd;\wNorm{0:\NumNeg}}]{ \gradP \log \pUnnorm(\y_{\selectInd}) }}  \nonumber \\
    &= \expect[\pNoise(\y_{1:\NumNeg})]{\sum_{\numNeg=0}^{\NumNeg}  \wNorm{\numNeg} \gradP \log \pUnnorm(\y_\numNeg)}.
\end{align}
Approximating the expected value with a single Monte Carlo sample $\y_{1:\NumNeg} \sim \pNoise(\cdot)$ and plugging in the expression for the weights $\wNorm{\numNeg}$ when evaluating \cref{eq:cd:grad}, we recover \cref{eq:rank_nce:grad}.

\subsection{CNCE criterion derivation}
\label{app:cnce-criterion}
The CNCE criterion is a proxy-criterion based on a binary classification problem with true sample $\y_0 \sim \pData(\cdot)$ and noisy sample $\y_1 \sim \pNoise(\cdot \mid \y_0)$. We forget the origin of $\y_0, \y_1$ and introduce the latent class variable $\bernoullivar \in \{0, 1\}$ according to
\begin{align}
    \label{eq:cond_nce:cond_prob:simple}
    \pModel(\y_0, \y_1 \mid \bernoullivar ) =
    \begin{cases}
        \pModel(\y_0) \pNoise(\y_1 \mid \y_0), &\text{if } \bernoullivar = 0, \\
        \pModel(\y_1) \pNoise(\y_0 \mid \y_1), &\text{if } \bernoullivar = 1.
    \end{cases}
\end{align}
Let $p(z=0) = \eta$. %
The posterior of $\bernoullivar$ follows as
\begin{align}
    \label{eq:cond_nce:cond_prob:simple:z_1}
    \pModel(\bernoullivar = 0 \mid \y_0, \y_1)
    &= \frac{\pModel(\y_0, \y_1 \mid \bernoullivar=0) \eta}{\pModel(\y_0, \y_1 \mid \bernoullivar=0) )\eta + \pModel(\y_0, \y_1 \mid \bernoullivar=1) )(1 - \eta)} \nonumber \\
    &= \frac{\pModel(\y_0) \pNoise(\y_1 \mid \y_0) \eta }{\pModel(\y_0) \pNoise(\y_1 \mid \y_0) \eta + \pModel(\y_1) \pNoise(\y_0 \mid \y_1) (1-\eta) } \nonumber \\
    &= \frac{\pUnnorm(\y_0) \pNoise(\y_1 \mid \y_0) \eta }{\pUnnorm(\y_0) \pNoise(\y_1 \mid \y_0) \eta + \pUnnorm(\y_1) \pNoise(\y_0 \mid \y_1)(1-\eta) } \nonumber \\
    &= \frac{1}{1 + \displaystyle\frac{\w(\y_1 \mid \y_0) (1-\eta)}{\w(\y_0 \mid \y_1) \eta}}.
\end{align}
Note that the normalisation constant of $\pModel$ cancels, so that we can use the unnormalised model $\pUnnorm$ directly to calculate the posterior. In accordance with \citet{cond_nce}, we assume a uniform prior on $\bernoullivar$, i.e. $\eta=\frac{1}{2}$, 
as well as average over $\NumNeg$ noisy samples,
which yields the CNCE criterion in \cref{eq:cond_nce:crit}.
\subsection{Gradient of CNCE criterion}
\label{app:cnce-gradient}
We derive the gradient of the CNCE criterion in \cref{eq:cond_nce:crit}:

\begin{align}
  \gradP{} \condLossFn(\param, \y_{0:\NumNeg}) 
  &=   \frac{1}{\NumNeg}\sum_{\numNeg=1}^{\NumNeg} \gradP{}\log \left( 1 + \frac{\w(\x_{\numNeg} \mid \x_0)}{\w(\x_0 \mid \x_{\numNeg})} \right)\nonumber \\
   &= - \frac{1}{\NumNeg}\sum_{\numNeg=1}^{\NumNeg}\gradP{} \log \left( \frac{\pUnnorm(\y_0) \pNoise(\y_{\numNeg} \mid \y_0)  }{\pUnnorm(\y_0) \pNoise(\y_{\numNeg} \mid \y_0)  + \pUnnorm(\y_{\numNeg}) \pNoise(\y_0 \mid \y_{\numNeg}) } \right)\nonumber \\
  &=  \frac{1}{\NumNeg}\sum_{\numNeg=1}^{\NumNeg} \big(-\gradP{} \log  \left(\pUnnorm(\y_0) \pNoise(\y_{\numNeg} \mid \y_0)\right)  +  \gradP{} \log \big(\pUnnorm(\y_0) \pNoise(\y_{\numNeg} \mid \y_0) \nonumber \nonumber \\ & \eqspace  + \pUnnorm(\y_{\numNeg}) \pNoise(\y_0 \mid \y_{\numNeg}) \big) \big) \nonumber \\
  &=  - \gradP{} \log  \pUnnorm(\y_0)  +  \frac{1}{\NumNeg}\sum_{\numNeg=1}^{\NumNeg}\big( \gradP{} \log \big(\pUnnorm(\y_0) \pNoise(\y_{\numNeg} \mid \y_0) \nonumber \\ & \eqspace  + \pUnnorm(\y_{\numNeg}) \pNoise(\y_0 \mid \y_{\numNeg}) \big) \big) 
\end{align}
Let $\Zz = \pUnnorm(\y_0) \pNoise(\y_{\numNeg} \mid \y_0)  + \pUnnorm(\y_{\numNeg})\pNoise(\y_0 \mid \y_{\numNeg})$, then
\begin{align}
\gradP{} \log \Zz 
&= \frac{\gradP{} \Zz}{ \Zz}
= \frac{\pNoise(\y_{\numNeg} \mid \y_0) \gradP{} \pUnnorm(\y_0) + \pNoise(\y_0 \mid \y_{\numNeg}) \gradP{} \pUnnorm(\y_j)}{ \pUnnorm(\y_0) \pNoise(\y_{\numNeg} \mid \y_0) + \pUnnorm(\y_{\numNeg}) \pNoise(\y_0 \mid \y_{\numNeg}) } \nonumber \\
&= \frac{\pNoise(\y_{\numNeg} \mid \y_0) \pUnnorm(\y_0) \gradP{} \log \pUnnorm(\y_0) + \pNoise(\y_0 \mid \y_{\numNeg}) \pUnnorm(\y_{\numNeg}) \gradP{} \log \pUnnorm(\y_{\numNeg})}{ \pUnnorm(\y_0) \pNoise(\y_{\numNeg} \mid \y_0) + \pUnnorm(\y_{\numNeg}) \pNoise(\y_0 \mid \y_{\numNeg}) } \nonumber \\
&= 
\wNormCond{0}{j}
\gradP{} \log \pUnnorm(\y_0) + 
\wNormCond{j}{0}
\gradP{} \log \pUnnorm(\y_{\numNeg}) \nonumber \\
&= 
(1-\wNormCond{j}{0}) 
\gradP{} \log \pUnnorm(\y_0) + 
\wNormCond{j}{0}
\gradP{} \log \pUnnorm(\y_{\numNeg}).
\end{align}
%

\subsection{Proof of \cref{prop:cd_cnce:cd_connection}}
\label{app:proof-cd-cnce}
To show that CNCE is equivalent to using CD-1 together with the kernel given by \cref{alg:cd_cnce:cnce_kernel}, we calculate the expectation with respect to $\condKernel$, initialising at $\y_0$
\begin{align}
    \label{eq:cnce:cd_derivation}
    &\expect[\condKernel(\y' \vert \y_0)]{\gradP{} \log \pUnnorm(\y')}
    \nonumber \\ & \eqspace
    = \mathbb{E}_{\pNoise(\y_{1} \mid \y_{0})} \big[ (1 - \wNormCond{1}{0})\gradP{} \log \pUnnorm(\y_0) 
     + \wNormCond{1}{0} \gradP{} \log \pUnnorm(\y_1) \big]. 
\end{align}
We recover \cref{eq:cond_nce:grad} by approximating the expectation with an average over samples $\y_{j} \sim \pNoise(\cdot \mid \y_{0})$, $j=1,\,\dots,\,\NumNeg$ and by plugging the result into \cref{eq:cd:grad}.
\subsection{Proofs of \cref{thm:q:rank:unbiased,thm:q:cond:unbiased}}
\label{app:proof:gradient-rnce-cnce}
With $q = \pModel$ we have uniform weights 
$\wNorm{\numNeg} = \frac{1}{\NumNeg+1}, \numNeg = 0, \dots, \NumNeg$ (from \cref{eq:rank_nce:grad}).
From \cref{eq:rank_nce:grad}, we then have
\begin{align}
  \label{eq:rank_crit:grad:model_prop}
  \gradP{} \rankLossFn(\param)
  &=- \gradP{} \log \pUnnorm(\y_0) + \sum_{\numNeg=0}^{\NumNeg} \frac{1}{\NumNeg + 1} \gradP{} \log \pUnnorm(\y_{\numNeg}) \nonumber \\
  &=- \frac{J}{\NumNeg + 1} \gradP{} \log \pUnnorm(\y_0) + \frac{1}{\NumNeg + 1} \sum_{\numNeg=1}^{\NumNeg} \gradP{} \log \pUnnorm(\y_{\numNeg}).
\end{align}
Taking the expectation \wrt $\pNoise(\y_{1:\NumNeg})=\pModel(\y_{1:\NumNeg})$ 
\begin{align}
\expect[\pNoise(\y_{1:\NumNeg})]{\gradP{} \rankLossFn(\param)}
&= - \frac{J}{\NumNeg + 1} \gradP{} \log \pUnnorm(\y_0) 
+ \frac{1}{\NumNeg + 1} \sum_{\numNeg=1}^{\NumNeg} \expect[\pModel(\y_\numNeg)]{\gradP{} \log \pUnnorm(\y_\numNeg)} \nonumber \\
&= \frac{J}{\NumNeg + 1} \left( - \gradP{} \log \pUnnorm(\y_0) + \gradP{} \log \partFn \right) \nonumber \\
&= \frac{\NumNeg}{\NumNeg + 1} \left( - \gradP{} \log \pModel(\y_0) \right),
\end{align}
which proves \cref{thm:q:rank:unbiased}.

Analogous to the above proof, 
with $q(\cdot \mid \x_0) = \pModel(\cdot)$ we have uniform weights 
$\wNorm{}(\y_0 \mid \y_{\numNeg}) = \wNorm{}(\y_{\numNeg} \mid \y_0) = \frac{1}{2}$, $\numNeg = 1, \ldots, \NumNeg$.
From \cref{eq:cond_nce:grad}, we then have
\begin{align}
  \gradP{} \condLossFn(\param)
  &= - \gradP{}\log \pUnnorm(\y_0) + \frac{1}{\NumNeg} \sum_{\numNeg=1}^{\NumNeg} \left( \frac{1}{2} \gradP{} \log \pUnnorm(\y_{0}) + \frac{1}{2} \gradP{} \log \pUnnorm(\y_{\numNeg}) \right) \nonumber \\
  &= - \frac{1}{2} \gradP{}\log \pUnnorm(\y_0) + \frac{1}{2 \NumNeg} \sum_{\numNeg=1}^{\NumNeg} \gradP{} \log \pUnnorm(\y_{\numNeg}).
\end{align}
Taking the expectation \wrt $\pNoise(\y_{1:\NumNeg} \mid \y_0)=\pModel(\y_{1:\NumNeg})$ 
\begin{align}
  \expect[\pNoise(\y_{1:\NumNeg} \mid \y_0)] {\gradP{} \condLossFn(\param)}
  &= - \frac{1}{2} \gradP{}\log \pUnnorm(\y_0) + \frac{1}{2 \NumNeg} \sum_{\numNeg=1}^{\NumNeg} \expect[\pModel(\y_\numNeg)] { \gradP{} \log \pUnnorm(\y_{\numNeg})} \nonumber \\
  &= \frac{1}{2} \left( - \gradP{}\log \pUnnorm(\y_0) + \gradP{} \log \partFn \right) \nonumber \\
  &= \frac{1}{2} \left( - \gradP{}\log \pModel(\y_0)\right),
\end{align}
which proves \cref{thm:q:cond:unbiased}.
\subsection{Proof of \cref{thm:unbiased_cis:prop_loss}}
\label{app:proof-adaptive-unbiased}
From \cref{eq:prop_distr:kl_crit:grad} we have 
\begin{align}
    \grad \lossFn(\varphi) &= \expect[\pModel(\y)]{ - \grad \log \pNoiseF(\y)}
    \approx
    \grad \lossFnHat(\varphi) = - \sum_{\numNeg=0}^\NumNeg \wNorm{\numNeg} \gradF \log \pNoiseF(\y_\numNeg) \nonumber \\
    &= - \frac{1}{\NumNeg+1} \sum_{\numNeg=0}^\NumNeg \frac{\wabbrv{\numNeg} \gradF \log \pNoiseF(\y_\numNeg)}{\partFnHatCis}.
\end{align}
Then
\begin{align}
    \expect[\pModel(\y_0) \pNoiseF(\yNegSeq{1})]{\grad \lossFnHat(\varphi)}
    &= \expect[\p(z, \yNegSeq{0})]{\grad \lossFnHat(\varphi)} \nonumber \\ 
    &= - \frac{1}{\NumNeg+1} \sum_{\numNeg=0}^\NumNeg \expect[\p(z, \yNegSeq{0})]{ \frac{1}{\partFnHatCis} \wabbrv{\numNeg} \gradF \log \pNoiseF(\y_\numNeg) }.
\end{align}
From \cref{thm:unbiased_cis:general} we have
\begin{align}
    \expect[\p(z, \yNegSeq{0})]{ \frac{\wabbrv{\numNeg} \gradF \log \pNoiseF(\y_\numNeg)}{\partFnHatCis}}
    &= \frac{1}{\partFn} \expect[\pNoiseF(\y)]{ \w(\y) \gradF \log \pNoiseF(\y) } \nonumber \\ 
    &= \expect[\pNoiseF(\y)]{ \frac{1}{\partFn} \w(\y) \gradF \log \pNoiseF(\y) } \nonumber \\
    &= \int \pNoiseF(\y) \frac{1}{\partFn} \w(\y) \gradF \log \pNoiseF(\y) \d \y \nonumber \\
    &= \int \pNoiseF(\y) \frac{1}{\partFn} \frac{\pUnnorm(\y)}{\pNoiseF(\y)} \gradF \log \pNoiseF(\y)\d \y \nonumber \\
    &= \int \pModel(\y) \gradF \log \pNoiseF(\y)\d \y \nonumber \\ &
    = \expect[\pModel(\y)]{\gradF \log \pNoiseF(\y)} = - \gradF \lossFn(\varphi).
\end{align}
Then
\begin{align}
    \expect[\pModel(\y_0) \pNoiseF(\yNegSeq{1})]{\grad \lossFnHat(\varphi)}
    &= \frac{1}{\NumNeg+1} \sum_{\numNeg=0}^\NumNeg \gradF \lossFn(\varphi),
     = \gradF \lossFn(\varphi),
\end{align}
which concludes the proof.
\section{Extensions of NCE}
\label{app:nce_extensions}
In the main paper, we outlined a new adaptive proposal strategy, persistent NCE, an MH-variant of CNCE as well as an SMC variant of RNCE, as extensions of the NCE criteria. Here, we detail one additional extension based on CD, namely that of taking several MCMC steps in the CD kernel. We also give further details on the SMC variant of RNCE (SMC-RNCE).  

\subsection{NCE with multiple MCMC steps}
\label{app:sec:multi-step}
In the light of interpreting RNCE and  CNCE as special cases of contrastive divergence, a natural extension of these criteria is that of taking several MCMC steps in the kernel. This, with the hope that it will improve convergence of the algorithm.  We outline the procedure for RNCE with $\totMcmcSteps$ MCMC steps, using the kernel in \cref{alg:cd_rnce:rnce_kernel}. At each MCMC step $\mcmcStep = 1, \ldots, \totMcmcSteps$, we sample $\y_{1:\NumNeg}^{(\mcmcStep)}$ and condition on $\y_0^{(\mcmcStep)} = \y_{\selectInd}^{(\mcmcStep - 1)}$, with $\y_{\selectInd}^{(\mcmcStep - 1)}$ being the sampled output from the previous step.
Marginalising over the index variables, $\selectInd$, we estimate the second term in \cref{eq:cd:grad} as
\begin{align}
    \label{eq:cd_rnce_multi_steps:grad}
    &\expect[\rankKernel(\y' \vert \y_0)]{ \gradP \log \pUnnorm(\y') } \approx \frac{1}{\totMcmcSteps} \sum_{\mcmcStep = 1}^{\totMcmcSteps} \sum_{\numNeg=0}^{\NumNeg} \wNorm{\numNeg}^{(\mcmcStep)} \gradP{\log \pUnnorm(\y_{\numNeg}^{(\mcmcStep)})}.
\end{align}
Hence, we estimate the expected gradient as an average over the estimates obtained at each step of the kernel. Note that the weight normalisation, \cref{eq:rank_nce:grad}, is performed independently at each step $\mcmcStep$, using only samples involved in that particular step. The procedure would be similar for CNCE, but instead employing the kernel outlined in \cref{alg:cd_cnce:cnce_kernel}.

\subsection{Sequential Monte Carlo RNCE}
Interpreting RNCE as CD-1 with a kernel based on CIS, we propose an extension to RNCE, where the CIS kernel is replaced by a kernel based on conditional Sequential Monte Carlo (SMC), see e.g. \citep{elements_of_smc}, referred to by SMC-RNCE. Here, we give details on the CSMC algorithm. The CSMC kernel is outlined in \cref{alg:csmc:csmc_kernel}. 

SMC, in general, tries to address the issue of weight degeneracy sometimes observed in IS, see e.g. \citep{elements_of_smc}, by solving the inference problem recursively.
Assume that the model density factorises as 
\begin{align}
      p_\theta(\y) = \frac{1}{\partFn}\prod_{\xdim=1}^{\Xdim} \pUnnorm(\y_{\xdim}\mid \y_{1:(\xdim-1)}),
     \label{eq:nce_ext:ar_model}
\end{align}
for a given ordering of the $\Xdim$ features $\y_{\xdim}$, $\xdim=1, \ldots, \Xdim$, and where $\y_{1:(\xdim-1)} = [\y_{1}, \ldots, \y_{\xdim-1}]$. 
In this case, SMC can make use of the autoregressive structure to recursively draw samples from the proposal distribution and by adapting said distribution based on the previously drawn samples. 

In Conditional SMC (CSMC), similar to CIS, we condition on a sample $\x_0$ (a data sample in SMC-RNCE), which is set deterministically in the SMC algorithm. Following the notation in \cref{eq:nce_ext:ar_model}, CSMC iterates over all features starting at $\xdim=1$ and ending at $\xdim=\Xdim$. At step $\xdim$, samples $\y^{(\numNeg)}_{1:\xdim}$, $\numNeg=1, \ldots, \NumNeg$, are drawn from the proposal
\begin{align}
    \pNoise(\y_{1:\xdim}) = \pNoise(\y_{\xdim}\mid \y_{1:(\xdim-1)})\sum_{\numNeg=0}^{\NumNeg}\wNorm{\numNeg, \xdim-1}\delta_{\y^{(\numNeg)}_{1:(\xdim-1)}}(\y_{1:(\xdim-1)}),
    \label{eq:app:nce_ext:csmc_proposal}
\end{align}
with $\delta_{\y^{(\numNeg)}_{1:(\xdim-1)}}(\cdot)$ the Dirac delta distribution at the previously drawn sample $\y^{(\numNeg)}_{1:(\xdim-1)}$ (or the conditioning sample for the case $\numNeg=0$). The weights at step $\xdim$ are calculated as
\begin{align}
      &\wNorm{\numNeg, \xdim} = \frac{ w_\theta(\y^{(\numNeg)}_{\xdim})}{\sum_{\ell=0}^{\NumNeg}  w_{\theta}(\y^{(\ell)}_{\xdim})},     \label{eq:app:nce_ext:weight_norm}
\end{align}
with, for a model that factorises according to \cref{eq:nce_ext:ar_model}, 
\begin{align}
     &w_\theta(\y^{(\numNeg)}_{\xdim}) 
     = \frac{\pUnnorm( \y^{(\numNeg)}_{\xdim} \mid  \y^{(\numNeg)}_{1:(\xdim-1)})}{\pNoise( \y^{(\numNeg)}_{\xdim} \mid  \y^{(\numNeg)}_{1:(\xdim-1)})}
    \label{eq:app:nce_ext:weight_unnorm}.
\end{align}

The CSMC estimate of the normalisation constant, $\partFn$, is 
\begin{align}
    \partFnHatCsmc = \prod_{\xdim=1}^{\Xdim}\frac{1}{\NumNeg +1}\sum_{\numNeg=0}^{\NumNeg} w_\theta(\y^{(\numNeg)}_{\xdim}).
    \label{eq:app:csmc_zhat}
\end{align}

One issue that can arise in SMC is so-called path degeneracy, see e.g. (Naesseth et al.,
2019). To alleviate this issue, we can use \textit{adaptive resampling} and sample $\y^{(\numNeg)}_{1:(\xdim-1)}$, $\numNeg=1, \ldots, \NumNeg$, at step $\xdim$ only if the effective sample size ($\text{ESS}_{\xdim-1}$) goes below $(\NumNeg + 1) / 2$, and otherwise keep the corresponding samples from the last iteration. The effective sample size is calculated as
\begin{align*}
    \text{ESS}_{\xdim} = \frac{1}{\sum_{\numNeg=0}^{\NumNeg} \wNorm{\numNeg, \xdim}^2}.
\end{align*}
In case we do not resample $\y^{(\numNeg)}_{1:(\xdim-1)}$, we account for this by calculating the weights according to
\begin{align}
     &w_\theta(\y^{(\numNeg)}_{\xdim}) = \frac{\wNorm{\numNeg, \xdim-1}}{1/(\NumNeg + 1)}\frac{\pUnnorm( \y^{(\numNeg)}_{\xdim} \mid  \y^{(\numNeg)}_{1:(\xdim-1)})}{\pNoise( \y^{(\numNeg)}_{\xdim} \mid  \y^{(\numNeg)}_{1:(\xdim-1)})}
     \label{eq:app:nce_ext:weight_unnorm_resamp}.
\end{align}
These adapted weights are then used in the CSMC estimate of the normalisation constant (\cref{eq:app:csmc_zhat}).

\begin{algorithm}[H]
    \caption{CSMC kernel}
    \textbf{Input:} $\x_0$\\
    \textbf{for} $\xdim=1$ \textbf{to} $\Xdim$ \textbf{do}\\
    \hspace*{0.5cm} \textbf{for} $\numNeg=1$ \textbf{to} $\NumNeg$ \textbf{do}
        \begin{enumerate}[leftmargin=1.5cm, topsep=0.2em, itemsep=0.05em]
            \item \textbf{if} $\xdim=1$\\
            \hspace*{0.5cm}Set $\y^{(\numNeg)}_{1:(\xdim-1)}\eqDef \emptyset$\\
            \textbf{else}\\
            \hspace*{0.5cm}Sample $\selectInd \sim \categorical{[\wNorm{0,\xdim-1}, \ldots, \wNorm{\NumNeg,\xdim-1}]}$, set $\y^{(\numNeg)}_{1:(\xdim-1)}\eqDef \y^{(\selectInd)}_{1:(\xdim-1)}$
            \item Sample $\y^{(\numNeg)}_{\xdim} \sim \pNoise(\cdot \mid \y^{(\numNeg)}_{1:(\xdim-1)})$
            \item Calculate weight $w_\theta(\y^{(\numNeg)}_{\xdim})$, using \cref{eq:app:nce_ext:weight_unnorm}
        \end{enumerate}
     \begin{enumerate}[leftmargin=1cm, topsep=0.2em, itemsep=0.05em]
        \setcounter{enumi}{3}
        \item Calculate weights $w_\theta(\y^{(0)}_{\xdim})$,  $\wNorm{\numNeg,\xdim}$, $\numNeg =0, \ldots, \NumNeg$, using \cref{eq:app:nce_ext:weight_unnorm,eq:app:nce_ext:weight_norm}
    \end{enumerate}
 \begin{enumerate}[leftmargin=0.5cm, topsep=0.2em, itemsep=0.05em]
     \setcounter{enumi}{5}
    \item Sample $\selectInd \sim \categorical{[\wNorm{0,\Xdim}, \ldots, \wNorm{\NumNeg,\Xdim}]}$
     \item Return $\y' = \y^{(\selectInd)}_{1:\Xdim}$
\end{enumerate}
\label{alg:csmc:csmc_kernel}
\end{algorithm}

\section{Comparison with Yair and Michaeli (2021)}
\newcommand{\ap}{\alpha_\param}
Most similar to our contribution is the work by \citet{cd_time_rev}.
They show that CD-1 can be derived from CNCE, while we show that CNCE (as well as RNCE) can be derived as special cases of CD-1.
Both results give valuable insight to two important families of estimation methods.
However, our two different approaches lead to some crucial distinctions, both conceptual and theoretical, which we detail below.

\citet{cd_time_rev} use the original derivation of the CD gradient, starting at another objective function than the log-likelihood, see \citep{contr_div_hinton}. 
From there, they argue that this derivation is flawed, since it assumes that an intractable term can be neglected.
Deriving CD-1 from CNCE is therefore more principled, they claim. 
Here, we take the opposite view.
We view the CD gradient in \cref{eq:cd:grad} as a straightforward MCMC approximation of the log-likelihood gradient in \cref{eq:unnorm_model:grad}.
This view is common and leads to exactly the same gradient expression, see \eg \citep{cd_wormholes,mcmc_mle}.
NCE on the other hand, is derived by introducing a proxy-criterion without any apparent connection to standard ML estimation.
Therefore, we claim that it is more useful to formulate the NCE methods in terms of their connection to ML.

The theoretical differences stem from the way \cite{cd_time_rev} derive their connection.
Specifically, they rewrite the gradient of the CNCE criterion on the form
\begin{subequations}
\begin{align}
  \gradP{} \condLossFn(\param) 
  &= \expect[\y_0 \sim \pData, \y_1 \mid \y_0 \sim \pNoise]{ \ap(\y_0, \y_1) \left( - \gradP{} \log \pModel(\y_0) + \gradP{} \log \pModel(\y_1) \right) } \\
  \ap(\y_0, \y_1) &= \left( 1 + \frac{\pModel(\y_0) \pNoise(\y_1 \mid \y_0)}{\pModel(\y_1) \pNoise(\y_0 \mid \y_1)} \right)^{-1}.
\end{align}
\end{subequations}
If $\pNoise$ is chosen to be the transition probability of a reversible Markov chain, then the detailed balance condition is fulfilled and $\ap(\y_0, \y_1) = 1/2,\ \forall \y_0, \y_1$, which means that the gradient of the CNCE criterion is proportional to the CD-1 gradient.
Note that the derivation holds only for proposal (or noise) distributions which satisfy this property.

Going in the opposite direction, we start from the CD gradient estimate in \cref{eq:cd:grad} and derive both CNCE and RNCE as special cases of CD-1.
We establish these links without any restrictions on the proposal distribution $\pNoise$.
By deriving CNCE from CD-1, we are also able to discover that CNCE corresponds to CD with a well-known MH kernel, albeit with a sub-optimal acceptance probability.
This allows us to propose a theoretical improvement to CNCE (MH-CNCE), at virtually no cost.

Our interpretation of RNCE and CNCE as CD-1 also provides a strong argument for choosing $\pNoise$ similar to $\pModel$, rather than $\pData$.
\cite{cd_time_rev} are more ambiguous on this point and simultaneously claim that $\pNoise$ should not significantly deviate from $\pData$, while also requiring that $\pNoise$ depend on $\pModel$.

We also arrive at different ways of generalising the methods to CD-$k$, that is with multi-step sampling in the MCMC kernel.
the multi-step versions of RNCE and CNCE follows naturally by taking multiple steps in the respective MCMC kernels, resulting in CD-$k$, see \cref{app:sec:multi-step}. 
This avoids the need to introduce the "time-reversal classification task" used by \citet{cd_time_rev}.
Importantly, we do pair-wise comparison of the conditional sample and the newly proposed one at each step of the MCMC kernel, while \cite{cd_time_rev} compare the full MCMC chain to the reversed one for all $k$ steps at once.

\section{Additional experiments and experimental details}
We provide additional details on the experiments performed in the main article, as well as additional experimental results.
\label{app:exp}
\subsection{Choice of proposal distribution}
The model $\pModel$ is parameterised by a mean vector $\bm{\mu}_\param$ and a vector $\bm{s}_\param$,
such that $\Sigma_\param$ is a diagonal covariance matrix, where the diagonal is the elements of $\bm{s}_\param$ squared.
We initialise $\bm{\mu}_{\param} = \bm{4}$ and $\Sigma_{\param} = 2 \identityMatrix$.

The models are trained using $\NumSamples = 100$ samples from $\pData$ and $\NumNeg=10$ proposal samples from $\pNoise$ for every data point drawn from $\pData$.
The parameters are estimated using SGD with learning rate $\lr = 0.01 \sqrt{\batchSize}$, where $\batchSize = 32$ is the batch size.
The parameter vectors $\param$ and $\varphi$ are updated once every batch, using the same learning rate.
\subsection{MH variant and persistent CNCE}
We provide additional results, investigating the effect of changing the CNCE acceptance probability to the one of the standard Metropolis--Hastings (MH) algorithm as well as that of using persistence in CNCE and MH-CNCE (P-CNCE and P-MH-CNCE, respectively). We also give additional details on the ring model experiments performed in the main paper. 

We perform experiments using the ring model explained in \cref{sec:exp:ring_model} with $\NumSamples = 200, 1000$ and $\NumNeg=5, 10$. The proposal $\pNoise$ is the same as previously. Each experiment is repeated 100 times, each time with a new set of uniformly sampled model parameters $\mu \in \{5, 10\}$, $\sigma^2 \in \{0.3, 1.5\}$, with $\theta = \log(\sigma^{-2})$, and a new data set of $\NumSamples$ data points. Initial estimates of $\theta$ is sampled uniformly from the same interval as the true value. To the best of our ability, we follow the setup in \citep{cond_nce}, but select only two data set sizes, as well as run additional experiments with a smaller number of noise samples ($\NumNeg=5$). Moreover, we train our models using SGD, training each model for 50 epochs with a batch size of $\batchSize=20$. 

For improving stability of the persistent CNCE methods, we use a decaying learning rate. The learning rate is set as $\lr= \lr_{\text{base}} \cdot \sqrt{\batchSize}$ and is linearly decayed, starting at $\lr_{\text{base}}=0.01$ and ending at $\lr_{\text{base}}=0.001$. For P-CNCE as well as P-MH-CNCE, we run $\batchSize \cdot \NumNeg$ MCMC chains in parallel, as each data point $\y_0^i$ in a batch forms a total of $\NumNeg$ pairs $(\y_0, \y_{\numNeg})$, $\numNeg = 1, \ldots, \NumNeg$.

\begin{figure}[ht]
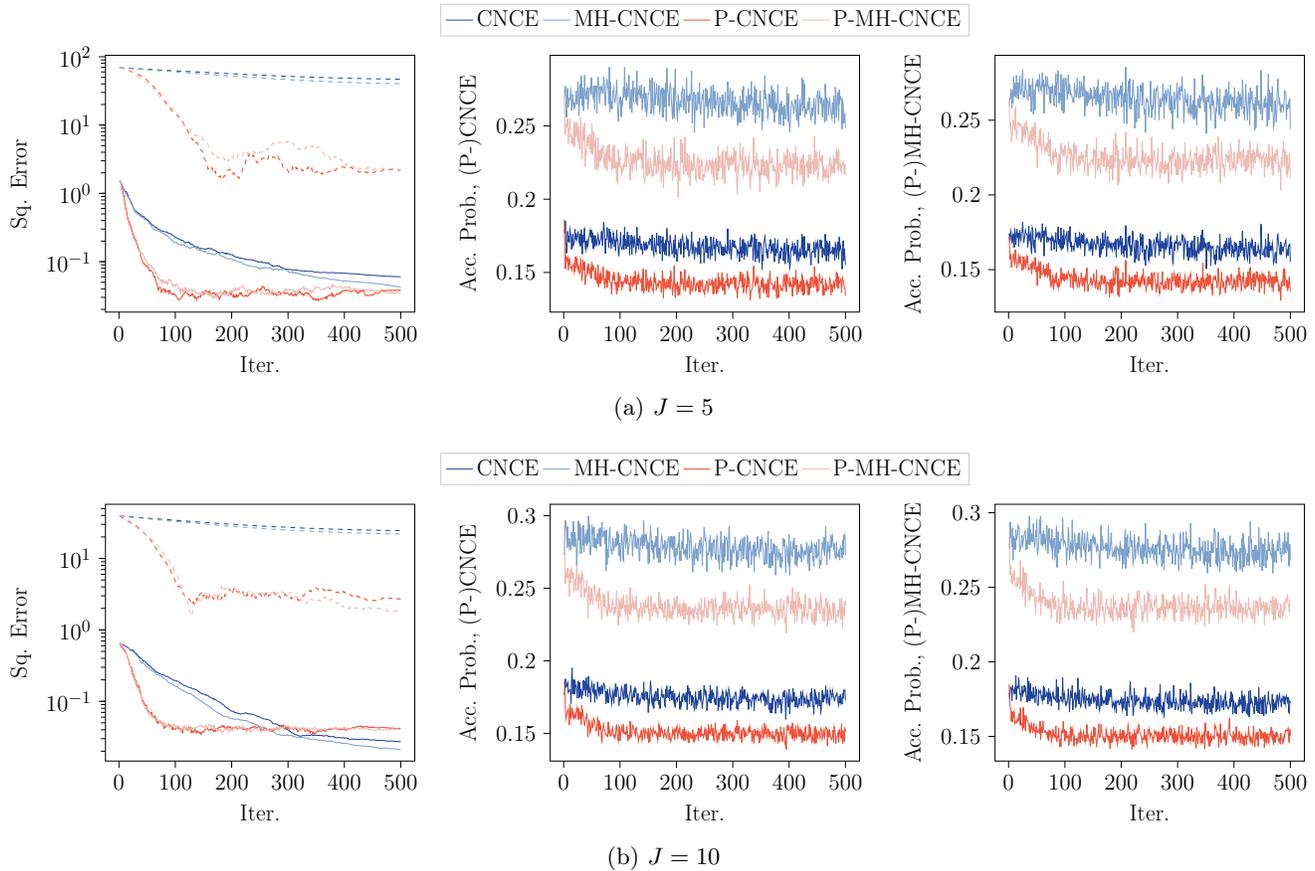

\vskip 0.2in
\begin{center}
\begin{subfigure}[]{\textwidth}
\centerline{\input{content/experiments/figs/cnce_acc_prob/cnce_acc_prob_res_num_samples_200_num_neg_samples_5.tex}}
\caption{$\NumNeg = 5$}
\end{subfigure}\vspace{1em}
\begin{subfigure}[]{\textwidth}
\centerline{\input{content/experiments/figs/cnce_acc_prob/cnce_acc_prob_res_num_samples_200_num_neg_samples_10.tex}}
\caption{$\NumNeg = 10$}
\end{subfigure}
\caption{Results for ring model experiments with $\NumSamples=200$ training data points. Results are reported over training iterations and as median (solid lines) and worst-case (dashed lines) estimated from 100 experiments.
Left: Squared parameter error of standard CNCE, CNCE with Metropolis--Hastings acceptance probability (MH-CNCE), persistent CNCE (P-CNCE) and persistent MH-CNCE (P-MH-CNCE). Middle: Acceptance probability of (P-)CNCE and (P-)MH-CNCE when training with (P-)CNCE. Right: Acceptance probability of CNCE and MH-CNCE when training with (P-)MH-CNCE.}
\label{fig:acc_prob_exp_additional_1}
\end{center}
\vskip -0.2in 
\end{figure}

Results are shown in \cref{fig:acc_prob_exp_additional_1,fig:acc_prob_exp_additional_2}. In addition to the median squared error of the estimated precision over iterations, we show the worst-case performance at each iteration. We also show the median acceptance probabilities obtained with all methods, calculated when training with CNCE and MH-CNCE for the standard algorithms, or P-CNCE and P-MH-CNCE for the persistent algorithms. 

The advantages of a higher acceptance probability as well as persistence is most evident for the smaller sample size. However, also for $N=1000$ we observe that both P-CNCE and P-MH-CNCE improves worst-case performance over standard CNCE and MH-CNCE. While the median performance of P-CNCE and P-MH-CNCE is worse, this still indicates that the methods have some robustness. The performance advantage of using the MH acceptance probability, compared to the one used in CNCE, is less evident when combined with persistence. 

\begin{figure}[ht]
\vskip 0.2in
\begin{center}
\begin{subfigure}[]{\textwidth}
\centerline{\input{content/experiments/figs/cnce_acc_prob/cnce_acc_prob_res_num_samples_1000_num_neg_samples_5.tex}}
\caption{$\NumNeg = 5$}
\end{subfigure}\vspace{1em}
\begin{subfigure}[]{\textwidth}
\centerline{\input{content/experiments/figs/cnce_acc_prob/cnce_acc_prob_res_num_samples_1000_num_neg_samples_10.tex}}
\caption{$\NumNeg = 10$}
\end{subfigure}
\caption{Results for ring model experiments with $\NumSamples=1000$ training data points. Results are reported over training iterations and as median (solid lines) and worst-case (dashed lines) estimated from 100 experiments.
Left: Squared parameter error of standard CNCE, CNCE with Metropolis--Hastings acceptance probability (MH-CNCE) and persistent CNCE (P-CNCE). Middle: Acceptance probability of CNCE and MH-CNCE when training with (P-)CNCE. Right: Acceptance probability of CNCE and MH-CNCE when training with (P-)MH-CNCE.}
\label{fig:acc_prob_exp_additional_2}
\end{center}
\vskip -0.2in 
\end{figure}
\subsection{Autoregressive EBM}\label{app:autoregr_ebm}

The experimental setup for the experiments performed with the autoregressive EBM (AR-EBM) are based on similar experiments performed by \cite{aem,ace}. Note that, while the experiments are based on previous work \citep{aem, ace}, we formulate the model distribution such that the conditional distributions in \cref{eq:nce_ext:ar_model} share a single normalisation constant, $\partFn$. In contrast, the model distributions in \citep{aem,ace} are formulated such that each one-dimensional conditional distribution has its own normalisation constant, which is also estimated separately from the other's.

For the experiments, we select reasonable hyperparameters based on values used in \citep{aem}. In some cases, we adapt the hyperparameters based on initial test runs using ML-IS or to reduce computational cost (see below). We also use the test runs, guided by \citet{aem}, to decide on a sufficient number of training iterations for each dataset. The hyperparameters are summarised in \cref{tab:autoregr_hps}. We found training with ML-IS on the Miniboone and BSDS300 datasets to be highly unstable. As we did not manage to find hyperparameters that stabilise training for this method combined with these datasets, we omit the corresponding results.

In all experiments, both the AR-EBM, $\pModel$, and the proposal network, $\pNoise_\varphi$, are fully-connected neural networks with residual connections, consisting of pre-activation residual blocks with two layers \citep{res_mapping}. For the AR-EBM we use four residual blocks with 128 hidden units for all datasets. For the proposal network, we use two residual blocks with 512 hidden units for the dataset of highest dimension (BSDS300) and two residual blocks with 256 hidden units for the remaining datasets. The size of the AR-EBM is directly taken from \cite{aem}, while we choose a smaller proposal network than used in \citep{aem}, to reduce computational cost. In both models, we use ReLU activations and apply dropout between the two layers of each residual block, using a dropout rate of 0.1, as a default. However, as we did not observe convergence of the loss on the Gas dataset using this setup, we exchange the activation function in the AR-EBM to Tanh and train without dropout for this dataset, following \cite{aem}. %
Moreover, for the smaller Miniboone dataset, we use a dropout rate of 0.5 as in \citep{aem}, to avoid overfitting. 
The structure of the proposal network is taken from \citep{ace}. The input to the network is the feature vector $\x_0$ with all but the observed features $x_{0, 1:(\xdim-1)}$ set to 0, and a one-hot mask indicating which features are observed (with zeros in the positions of unobserved features, and ones in the positions of observed features). The proposal network parameterises a Gaussian Mixture Model (GMM) with ten components. The GMM is used to evaluate the conditional density $\pNoise_\varphi(\y_{\xdim} \mid x_{1:(\xdim-1)})$ and to generate proposal/noise samples.
To avoid numerical issues, we set the standard deviation of each Gaussian to a minimum of $1\cdot 10^{-3}$. 

Apart from the parameters of the GMM, we follow previous experimental setups \citep{aem,ace} and let the proposal network output a context vector of length 64. This context vector, which can be interpreted as a latent representation of $x_{1:(\xdim-1)}$, is given as input to the AR-EBM as a means of sharing information between the two models. Given the context vector as well as the unobserved feature $x_{0, \xdim}$ as input, the AR-EBM directly predicts the unnormalised log density (the negative energy) $\log \pUnnorm(x_{\xdim} \mid x_{1:(\xdim-1)})$. As in \citep{aem}, we apply a softplus non-linearity to the output of the AR-EBM, such that the predicted density is upper bounded by 1.

Prior to training, data is pre-processed according to \cite{aem}. For the Power, Gas and Miniboone datasets we hold out 10\% of the data for testing, and split the remaining dataset into training and validation sets using a 90\%-10\% split. For Hepmass and BSDS300 we use pre-existing data splits. For Hepmass, this means that the test set makes up approximately one third (instead of $10 \%$) of the full dataset. For BSDS300, the test set instead consists of approximately $20 \%$ of the full dataset, and the remaining data is split into training and validation sets using a (approximately) $95\%-5\%$ split. Each dataset but BSDS300 is standardised and features that are discrete, highly correlated or with too many reoccurring values are removed. For the Power dataset, we additionally add uniform noise to the features, for the purpose of numerical stability. Moreover, instead of using the full images of the original BSDS300 dataset, we use the same data as in \citep{aem}, consisting of patches extracted from the original images. The final number of data dimensions, as well as the total size of each dataset are given in \cref{tab:autoregr_data}. 

The AR-EBM, $\pModel$, and proposal network, $\pNoise_\varphi$, are trained in parallel. We investigate several methods for training the AR-EBM (ML-IS, RNCE and SMC-RNCE) and train $\pNoise_\varphi$ using maximum likelihood as done in \citep{aem,ace}. Although we argue that $\pNoise_\varphi$ should resemble $\pModel$, we find that our proposed method for adapting $\pNoise_\varphi$ to the model distribution is not suitable for the particular setup where information is shared (through the context vector) between the AR-EBM and the proposal network. Having both $\pModel$ and $\pNoise_\varphi$ aiming at the same target (the data distribution), is beneficial in this case, as this updates the context vector in the same direction, while having different targets (the data vs. the model distribution) instead risks stalling training, as the proposal might have a negative impact on the AR-EBM through the context vector.

Just as in \citep{aem}, the models are trained using Adam optimisation \citep{adam} and a learning rate following a cosine annealing schedule. The initial learning rate is set to $5\cdot 10^{-4}$. We use a batch size of 512, with the exception for the Miniboone and BSDS300 datasets, where we, because of limitations in GPU memory, use a batch size of 128. The total number of training iterations used for each dataset is reported in \cref{tab:autoregr_hps}. In all cases, we use the first 5000 iterations as a warm-up phase, where we train only the proposal network and keep $\pModel$ fixed. For training the AR-EBM, all methods use $\NumNeg=20$ proposal/noise samples per observation. In SMC-RNCE, we use adaptive resampling, as explained in \cref{app:nce_extensions}. During training, we evaluate the model on a smaller part (10\%) of the validation set every 1,000 training iterations, and keep the one with the highest log-likelihood out of the evaluated (together with the corresponding proposal network). An exception is made for the Miniboone dataset, where we evaluate the models on the full validation set every 1,000 training iterations, because of the smaller number of observations in the dataset. 

We train the models on one GPU (NVIDIA GeForce RTX 3090, 24 GB). The total training time depends on the method used to train the AR-EBM as well as the dataset in question. For the Power, Gas and Hepmass datasets, training takes around 5-25 hours, while the corresponding numbers for the Miniboone and BSDS300 datasets are 15-40 and 40-120 hours, respectively.
We evaluate the log-likelihood over the test set, applying SMC to estimate the normalisation constant. We make a total of ten estimates, using $5\cdot 10^6$ particles in the SMC algorithm, and report the mean as well as standard error of the log-likelihood. The 2-Wasserstein distance is estimated with sampling. We randomly draw (with replacement) $1 \cdot 10^4$ samples from the test set, to represent the data distribution, and use SMC to draw an equal amount of samples from the model distribution. An exception is again made for the Miniboone dataset, where we draw only $2 \cdot 10^3$ from each distribution, because of the smaller size of the test set. We make a total of ten estimates, and report the mean as well as the standard error of the estimated Wasserstein distance. 

\begin{table}
  \caption{Dimension and total size of the datasets used in the AR-EBM experiments.}
  \label{tab:autoregr_data}
  \centering
  \begin{tabular}{lccccc} 
     & Power  & Gas  & Hepmass & Miniboone & BSDS300 \\
    \midrule
    Dimension ($D$) & 6 & 8 & 21 & 43 & 63 \\
    Size ($N$) & 2,049,280 & 1,052,065 & 525,123 & 36,488 & 1,300,000\\
    \bottomrule
  \end{tabular}
\end{table}

\begin{table}
  \caption{Hyperparameters used in the AR-EBM experiments.}
  \label{tab:autoregr_hps}
  \centering
  \begin{tabular}{lccccc} 
     & Power  & Gas  & Hepmass & Miniboone & BSDS300 \\
    \midrule
    Number of blocks, $\pModel$ & 4 & 4 & 4 & 4 & 4 \\
    Hidden dimension, $\pModel$ & 128 & 128 & 128 & 128 & 128 \\
    Activation function, $\pModel$  & ReLU & Tanh & ReLU & ReLU & ReLU \\
    Number of blocks, $\pNoise_\varphi$ & 2 & 2 & 2 & 2 & 2 \\
    Hidden dimension, $\pNoise_\varphi$ & 256 & 256 & 256 & 256 & 512 \\
    Activation function, $\pNoise_\varphi$  & ReLU & ReLU & ReLU & ReLU & ReLU \\
    Context dimension & 64 & 64 & 64 & 64 & 64 \\
    GMM components & 10 & 10 & 10 & 10 & 10 \\
    Negative samples, $\NumNeg$ & 20 & 20 & 20 & 20 & 20 \\
    Batch size & 512 & 512 & 512 & 128 & 128 \\
    Dropout rate ($\pModel$ and $\pNoise_\varphi$)  & 0.1 & 0.0 & 0.1 & 0.5 & 0.1 \\
    (Initial) learning rate & $5\cdot 10^{-4}$ & $5\cdot 10^{-4}$ & $5\cdot 10^{-4}$ & $5\cdot 10^{-4}$ & $5\cdot 10^{-4}$ \\ 
    Training iterations & $1\cdot 10^6$ & $6\cdot 10^5$ & $2\cdot 10^5$ & $3\cdot 10^5$ & $6 \cdot 10^5$ \\
    Warm-up iterations & 5000 & 5000 & 5000 & 5000 & 5000 \\
    \bottomrule
  \end{tabular}
\end{table}

\end{document}